\documentclass[acmtog]{acmart}%

\setcopyright{none}
\acmSubmissionID{145}

\usepackage{amsmath, amsthm, amsfonts, amssymb}
\usepackage{mathrsfs}
\usepackage{graphicx}
\usepackage{textcomp}
\usepackage{wrapfig}
\usepackage{subfig}
\usepackage{color}
\usepackage{xspace}
\usepackage{overpic}
\usepackage{subfig}
\usepackage{wrapfig}
\usepackage{enumitem}

\usepackage{color}
\definecolor{turquoise}{cmyk}{0.65,0,0.1,0.1}
\definecolor{purple}{rgb}{0.65,0,0.65}
\definecolor{dark_green}{rgb}{0, 0.5, 0}
\definecolor{orange}{rgb}{0.8, 0.6, 0.2}
\definecolor{red}{rgb}{0.8, 0.2, 0.2}
\definecolor{brown}{rgb}{0.5, 0.16, 0.16}

\renewcommand{\textrightarrow}{$\rightarrow$}

\begin{document}

\title{Predictive and Generative Neural Networks for Object Functionality}

\author{Ruizhen Hu}
\affiliation{%
	\department{College of Computer Science \& Software Engineering}
	\institution{Shenzhen University}
}
\email{ruizhen.hu@gmail.com}

\author{Zihao Yan}
\affiliation{%
	\institution{Shenzhen University}
}

\author{Jingwen Zhang}
\affiliation{%
	\institution{Shenzhen University}
}

\author{Oliver van Kaick}
\affiliation{%
	\institution{Carleton University}
}
\author{Ariel Shamir}
\affiliation{%
	\institution{The Interdisciplinary Center}
}
\author{Hao Zhang}
\affiliation{%
	\institution{Simon Fraser University}
}
\author{Hui Huang}
\authornote{Corresponding author: Hui Huang (hhzhiyan@gmail.com)}
\affiliation{%
	\department{College of Computer Science \& Software Engineering}
	\institution{Shenzhen University}
}

\renewcommand\shortauthors{R. Hu, Z. Yan, J. Zhang, O. van Kaick, A. Shamir, H. Zhang, and H. Huang}

\begin{teaserfigure}
  \includegraphics[width=\textwidth]{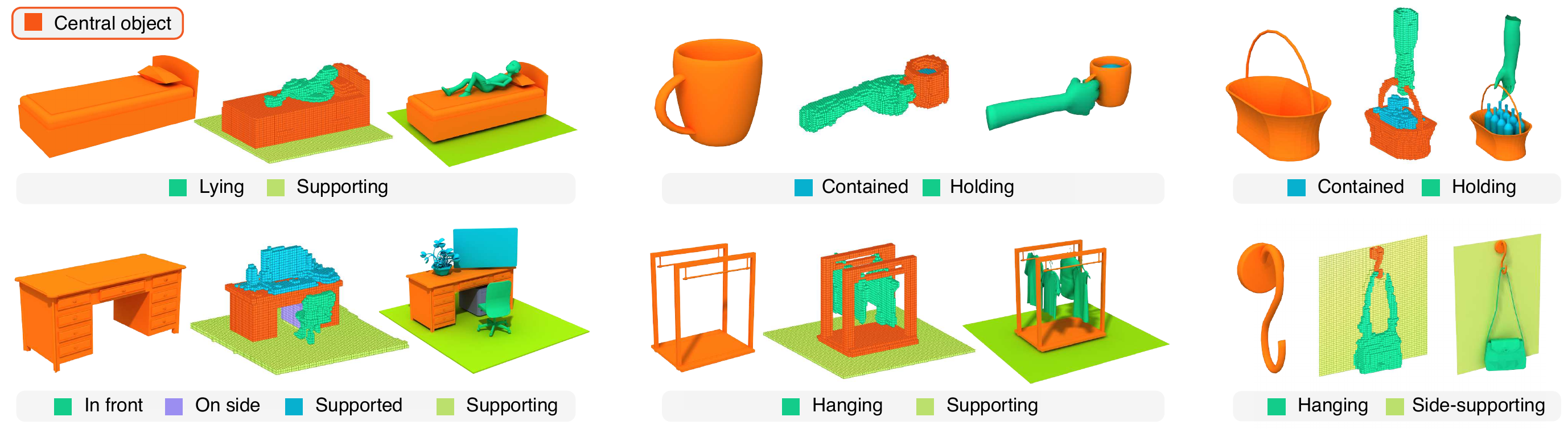}
  \caption{Given an object in isolation (left of each example), our generative network synthesizes scenes that demonstrate the functionality of the object in terms of interactions with surrounding objects (middle). Note the different types of functionalities appearing in the scenes generated by the network, involving interactions such as support, containment, and grasping. The scene is refined by replacing voxelized objects with higher resolution models (right). 
}
  \label{fig:teaser}
\end{teaserfigure}

\begin{abstract}
Humans can predict the functionality of an object even without any surroundings, since their knowledge and experience 
would allow them to ``hallucinate'' the interaction or usage scenarios involving the object. We develop {\em predictive\/} 
and {\em generative\/} deep convolutional neural networks to replicate this feat. Specifically, our work focuses on 
functionalities of man-made 3D objects characterized by human-object or object-object interactions. Our networks are
trained on a database of scene contexts, called {\em interaction contexts\/}, each consisting of a central object and
one or more surrounding objects, that represent object functionalities. Given a 3D object {\em in isolation\/},
our {\em functional similarity network\/} (fSIM-NET), a variation of the triplet network, is trained to predict the 
functionality of the object by inferring functionality-revealing interaction contexts. %
fSIM-NET is complemented by a generative network (iGEN-NET) and a segmentation network (iSEG-NET). iGEN-NET 
takes a single voxelized 3D object with a functionality label and synthesizes a voxelized
surround, i.e., the interaction context which visually demonstrates the corresponding functionality. iSEG-NET further 
separates the interacting objects into different groups according to their interaction types.

\end{abstract}




\setcopyright{acmcopyright}
\acmJournal{TOG}
\acmYear{2018}\acmVolume{37}\acmNumber{4}\acmArticle{151}\acmMonth{8} \acmDOI{10.1145/3197517.3201287}
\maketitle

\section{Introduction}
\label{sec:intro}

In recent years, functionality analysis of 3D shapes has gained attention as a means to understand and manipulate 3D environments. 
It has been argued that the recognition and categorization of object and scene data are mainly based on their functionality~\cite{stark91,greene16}. 
Even though functionalities of objects can be interpreted in many ways, most of them involve some form of {\em interactions\/} between two 
entities: one that provides the functionality and one that ``consumes'' it. 

Humans can predict the functionality of an object even without any surroundings, since their knowledge and experience would allow them to 
``hallucinate'' the interaction or usage scenarios involving the object. The main question we pose is whether a machine can replicate this feat, i.e., 
to predict the functionality of a 3D object given in isolation, possibly through an ``interaction hallucination'', and then be {\em generative\/}, 
i.e., to synthesize interactions that reflect the object's one or more functionalities (see Figure~\ref{fig:teaser}).
 
\begin{figure*}[!t]
    \centering
    \includegraphics[width=0.98\textwidth]{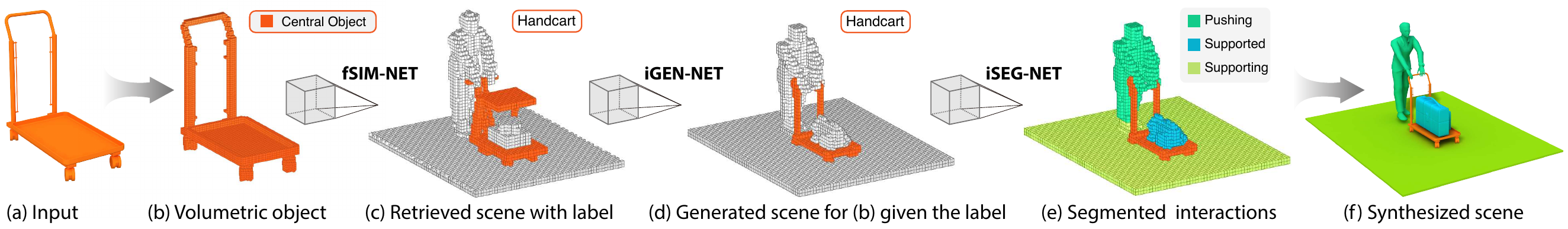}
\caption{Our complete framework for understanding object functionality using deep neural networks: Given a 3D object in isolation (a), we first transform it into a voxel representation (b). Then, we retrieve scenes with functionality most similar to that of the object (c), using our functional similarity network fSIM-NET. These retrieved scenes reveal functionalities in the form of interactions of the central object with surrounding objects, with associated functionality labels, e.g., ``Handcart''. Given such a functionality label from fSIM-NET, we can synthesize an interaction context (d) for the given 3D object (b) using our generative network iGEN-NET. Finally, we partition the interaction context into individual objects (e) using our segmentation network iSEG-NET, to enable further processing and analysis of the scene, such as replacing voxels with higher-resolution 3D models (f).}
\label{fig:overview}
\end{figure*}

Our work focuses on functionalities of man-made objects that are characterized by
human-object interactions (e.g., \emph{sitting on a chair} or \emph{pushing a cart}) or inter-object interactions (e.g.,
\emph{chairs next to a table} or \emph{books on a shelf}). 
The functionalities of an object can be revealed by a 3D scene that contains the object, designated as the {\em central object\/},
and one or more objects around it. These objects form a {\em scene context\/} within which 
one can study the central object's functionalities. It is the interactions between the central object and its surrounding objects that define
the functionalities. Similar to \cite{hu15}, we call the scene context an {\em interaction context\/}. 
To learn object functionalities, we take a data-driven approach and use a scene dataset, i.e., a set of interaction contexts.

Given an input 3D object {\em in isolation\/}, our goal is to train a model to infer interactions involving the object which reveal 
its functionalities, by learning from our scene dataset. We consider the inference task to have two facets: {\em prediction\/} 
and {\em synthesis\/}.
\vspace{-2pt}
\begin{enumerate}
\item In prediction, the key challenge is to learn a {\em space of interaction contexts\/} defined by {\em functional similarities\/}. 
Since we must deal with both isolated objects and scene data, we define the space of interaction contexts as a latent feature-space 
to which we map {\em both\/} objects and scenes. An isolated 3D object is mapped to a {\em distribution\/} over 
this space, allowing us to obtain interaction contexts which can help predict the object's multi-functionalities. 
\item In synthesis, we introduce a new problem to functionality analysis: to train a {\em generative\/} model which takes
a single 3D object with a functionality label as input and produces surrounding objects, i.e., the interaction context which reveals the corresponding functionality. %
\end{enumerate}

Defining functional similarity between objects or scenes is challenging since shape similarity does not suffice. 
Previous works on functionality analysis~\cite{savva16,hu16,pirk17} rely on specialized and hand-designed features such as
bisector surfaces or RAID~\cite{hu15,zhao14,guerrero16}. %
In contrast, our work learns the interaction context space via {\em metric learning\/} based on a novel artificial neural 
network: the {\em functional similarity network\/}, which we refer to as fSIM-NET.

Our fSIM-NET is a variation of the {\em triplet network}~\cite{wang14}, so that it becomes 
``{\em cross-domain\/}''. Specifically, each triplet input to the network consists of one 3D object in isolation, and two 
{\em scene instances\/}: one interaction context that positively reflects the functionality of the object and one
negative interaction context. The fSIM-NET learns a mapping of the inputs to the latent feature space based on 
functional similarity. %
It is trained by a novel triplet loss which pushes the mappings of the 3D object close to
that of the positive scene instance but away from the negative one.

For synthesis, we first introduce a {\em generative network\/}, coined iGEN-NET, which synthesizes an interaction context 
for a single input 3D object. Hence, after predicting functionality with the fSIM-NET, we can demonstrate it through a visual 
example. Note that an object-to-scene retrieval will not work for synthesizing such a scene, since the input object may be 
geometrically quite different from the central objects that exist in the training data. %
Lastly, we introduce a {\em segmentation network\/},
iSEG-NET, which takes the output synthesized by iGEN-NET and separates the interacting objects into different groups 
according to their interaction types. %

\vspace{-5pt}

\paragraph{Contributions.}
To the best of our knowledge, our work develops the first deep neural network for functionality analysis of 3D
objects. The key is to infer object functionalities by learning functional similarities via a novel triplet network and predicting interaction 
contexts for a single 3D object without relying on handcrafted features. The similarity network is complemented by a synthesis network,
followed by a segmentation network, to produce interaction contexts with segmented objects. The synthesis phase goes beyond inferring 
functionality labels or classes, it {\em substantiates\/} the acquired functional understanding. In conjunction, 
fSIM-NET, iGEN-NET, and iSEG-NET constitute an advanced framework for data-driven functional analysis of 3D objects; see
Figure~\ref{fig:overview}.

Our functional analysis framework consists of three separate, but streamlined, networks rather than a single unified network.
This facilitates preparation of training data and training of the networks. Specifically, the networks can be trained with example scenes that 
demonstrate a single functionality of each object, while the prediction and generation can involve multiple functionalities of an object.
The training of each network can be performed with loss functions targeted at subproblems of the framework and each individual network 
can be applied to serve different analysis tasks.

The three networks fSIM-NET, iGEN-NET, and iSEG-NET enable several applications, such as scene/object retrieval and object classification based 
on functionality, embedding of scenes and objects by a measure of functional similarity, and synthesis of interaction context scenes for individual objects. We also demonstrate in our evaluation that the similarity network provides a more accurate description of functionality when compared to previous work using hand-crafted descriptors~\cite{hu15,hu16}.

\section{Related work}
\label{sec:related}

\paragraph{Structural analysis of shapes.}
Part structures of shapes and functionality are intricately connected. Earlier works describe models of part structures that shapes 
with prescribed functionalities should possess. Such models can be manually defined~\cite{stark91} or partially learned from 
image data~\cite{pechuk08}. Structure-aware shape analyses indirectly discover functionalities by analyzing the shape of and 
relationships between shape parts~\cite{mitra13}, such as symmetry~\cite{wang11} and certain special support relations~\cite{zheng13}.

\vspace{-5pt}

\paragraph{Affordance analysis.}
Many works in computer graphics and vision %
analyze human-to-object interactions with the simulation of humanoid agents, to characterize object {\em affordance}. %
Grabner et al.~\shortcite{grabner11} explore the specific case of objects that can function as chairs, while more recent works are able to analyze a larger variety of humanoid interactions and object affordances~\cite{jiang13,kim14,zhu14,zhu15,wang17}. 
Going beyond the simulation of human poses, Savva et al.~\shortcite{savva14} track real human movements to learn {\em action maps} of scenes, encoding regions of an environment that can be used to perform certain tasks. Aside from object affordance, many functionalities also involve more general object-to-object interactions, which cannot be directly accounted for by human-object interactions.

\vspace{-5pt}

\paragraph{Encoding object-object interactions.}
To analyze object functionalities involving inter-object interactions, Zhao et al.~\shortcite{zhao14} introduce the Interaction Bisector Surface (IBS), which is a subset of the Voronoi diagram that encodes the space between any number of objects. Hu et al.~\shortcite{hu15} combine IBS with interaction regions into a hierarchical representation to constitute ICON, a geometric descriptor of interaction contexts. In subsequent work, Hu et al.~\shortcite{hu16} extend the ICON descriptor into a full model of functionality, which can be used to predict the functionality of a shape given in isolation.

The key limitation of these approaches is that the IBS and ICON descriptors are handcrafted encodings of interactions, using specific geometric constructions which may not encode all relevant information to describe 3D object functionalities. In contrast, our work introduces a data-driven approach for constructing such descriptors. We learn a feature encoding of object functionality using a convolutional neural network that maps voxelized objects to a latent space of interactions. Our cross-domain triplet network generalizes previous approaches to functionality analysis~\cite{zhao14,hu15,hu16} since it can measure functional similarities between scenes, and between objects and scenes, and it can predict the functionality of a 3D object given in isolation. Moreover, none of the previous works considered the problem of synthesizing contextual scenes of functional interactions from a single object.

\vspace{-5pt}

\paragraph{Interaction context vs.~ICON}
In both the work of Hu et al.~\shortcite{hu15} and ours, an interaction context is a 3D scene consisting of a central object and surrounding objects. However, the acronym ICON of Hu et al.~\shortcite{hu15} denotes a specific descriptor of interaction contexts, defined by
handcrafted features. In our work, the features are learned by a neural network. Throughout the paper, ICON is reserved to refer to the descriptor of Hu et al.~\shortcite{hu15}, while the term interaction context will be used as a generic reference to scene contexts.

\vspace{-5pt}

\paragraph{Scene synthesis.} 
Another line of work has focused on the generation of 3D indoor scenes. Fisher et al.~\shortcite{fisher12} learn a probabilistic model of object occurrence and arrangement from scene exemplars and Zhao et al.~\shortcite{zhao16} learn relationship templates for scene synthesis. More recent works consider interactions or human activities for the task. Savva et al.~\shortcite{savva16} learn a probabilistic model of human poses and spatial object configurations and apply the model to synthesize {\em interaction snapshots\/}. Ma et al.~\shortcite{ma16} learn a binding between human actions and object co-occurrences and placements from annotated images for action-driven 3D scene evolution. Fu et al~\shortcite{fu17} capture object arrangements and human activities with activity graphs learned from 2D floor plans and human positions. Differently from these works, we generate scenes with a deep neural network composed of mapping and decoder subnetworks. The relations between objects that are important for enabling certain functionalities are learned by the network directly from example scenes, and not handcrafted as in these works.

The interaction snapshots generated by Savva et al.~\shortcite{savva16} bear some resemblance to interaction contexts. An interaction snapshot consists of a human activity pose and one or more objects relevant to the activity (e.g., TV and sofa for watching TV). Their interaction snapshots are always human-centric, while our interaction contexts are more general. %
The most important distinction however, is that the inputs to their snapshot generation are terse yet explicit descriptions of one or more human activities in the form of verb-noun pairs, e.g., sit-chair $+$ use-laptop. In contrast, our generative network takes a single 3D object and synthesizes its surroundings based on a functionality label predicted by fSIM-NET.

\vspace{-5pt}

\paragraph{Neural networks for shape analysis and synthesis.}
Recently, research in deep neural networks has advanced significantly. %
In shape analysis, %
a few works learn a mapping from a high-dimensional volumetric grid into a lower-dimensional latent space, which can be sampled to synthesize new shapes~\cite{wu15,girdhar16,wu16}. Other works represent shapes as multi-view depth scans and learn a regression network for 3D shape completion~\cite{han17}. Alternative representations also include point sets~\cite{qi17}, processing directly on manifolds~\cite{masci15}, or a hierarchical representation suitable for the analysis and synthesis of man-made shapes~\cite{li17}. In our work, we also represent shapes and scenes as voxels, and develop the first predictive and generative convolutional neural networks for object functionality.

Triplet networks~\cite{wang14} are designed to offer distance metrics by learning an embedding function. 
In the classical setting, all three inputs to the network belong to the same domain,
while in fSIM-NET we compare two different representations: objects and scenes. 
Sung et al.~\shortcite{sung17} introduce a network for suggesting parts in model assembly, where retrieval and embedding subnetworks map parts and a partially-assembled model to a common space. By modeling part predictions as a probability distribution, their method can suggest multiple plausible parts for assembly. Similarly to this work, we also use subnetworks to map different representations to the same space, and model the prediction of functionality as a probability distribution. However, we address a completely different problem involving functionalities of objects and scenes, and model them as 3D volumes rather than point clouds.

\section{Overview}
\label{sec:overview}

The input and output of our analysis are 3D objects and scenes (Figure~\ref{fig:overview}). To use CNNs, we represent both objects and scene contexts 
by a cube of $64^3$ voxels. For individual objects, each voxel stores a binary value indicating whether the voxel is occupied 
by the object or not. For scenes, each voxel has three channels, where each channel holds a binary value indicating whether 
the voxel belongs to the central object, to an interacting object, or is empty. 

To learn our latent feature space for both objects and scenes, we design the fSIM-NET as a cross-domain
triplet network. Each triplet to the network consists of a {\em reference\/}, which is a 
single 3D \emph{object}, one positive example, which is a \emph{scene} context that positively reflects the functionality 
of the object, and one negative example, which is a random \emph{scene} context that is functionally dissimilar to the 
positive scene instances. The loss function of the network keeps a margin between two distances: the distance from the 
reference to the positive interaction context, and the distance from the reference to the negative interaction context. 
The fSIM-NET learns a mapping of all three inputs to the latent space of interaction contexts (see Figure~\ref{fig:network}).
However, %
we cannot use the same mapping for both objects and scenes. Instead, we 
learn {\em two\/} mappings: one maps scenes to the interaction context space, and the other maps an isolated object to a 
{\em distribution\/} over the same space. The distribution allows mapping one object to {\em multiple\/} regions in the 
space which may correspond to different functionalities exhibited by the same object (e.g., a wheeled chair can be sit on or pushed 
like a cart).

The object-to-interaction mapping by fSIM-NET is an essential starting point for understanding object functionalities.
In the synthesis phase, we develop two additional networks to complement the functionality prediction network.
In iGEN-NET, a generative convolution neural network, we take as input a 3D object in isolation in conjunction with a
functionality label inferred for the object using fSIM-NET. The output of the network is a voxelized 3D scene, i.e., an 
interaction context, which surrounds the input object with one or more objects to visually demonstrate the input functionality.

The iGEN-NET is a combination of an embedding network with a decoder and spatial transformer network, which synthesizes 
a scene and properly places the input 3D object into the scene. In contrast to previous indoor scene synthesis 
works~\cite{fisher12,savva16,ma16,fu17}, our network discovers the important shape features that reveal the 
interactions of the objects during the learning, rather than relying on handcrafted descriptors.

Lastly, the segmentation network, iSEG-NET, takes the output synthesized by iGEN-NET and separates the interacting objects 
into different groups according to their interaction types, enabling further analysis or post-processing of the objects, the scene, 
and the interactions involved.
The iSEG-NET is an encoder/decoder network that provides a labeling of the synthesized scene.

\section{fSIM-NET: functional similarity network}
\label{sec:method}

Our goal is to learn a distance metric $\mathcal{D}(x,Y)$ that reveals the dissimilarity between the functionality of an object $x$ given in isolation, and a central object provided with surrounding objects in a scene $Y$. The metric should enact the dissimilarity between the interactions that $x$ supports and the ones appearing in $Y$. In practice, to obtain this metric, we map objects and scenes to the space of interactions, and measure distances in this space. Thus, we reformulate our goal as learning two mapping functions: $E_{\text{obj}}$ for individual objects and $E_{\text{scn}}$ for scenes. Then, we can define $\mathcal{D}(x,Y) = \|E_{\text{obj}}(x) - E_{\text{scn}}(Y)\|_2$.  
The mapping functions should satisfy the requirement that scenes with similar interactions are close to each other in the mapping space, while scenes that support different interactions are far apart.
 
Similarly to previous works~\cite{wang14,lun15}, we learn the mapping functions from triplets that provide example instances of the metric. Specifically, our training set $\mathcal{T}$ is composed of triplets of the form $(x_i, Y_i^{+}, Y_i^{-})$, where $x_i$ is an object given in isolation, $Y_i^{+}$ is a positive example scene (a scene displaying the same functionality as $x_i$), and $Y_i^{-}$ is a negative example scene (not displaying the same functionality as $x_i$). Learning a meaningful metric can then be posed as learning $E_{\text{obj}}$ and $E_{\text{scn}}$ so that $\|E_{\text{obj}}(x_i) - E_{\text{scn}}(Y_i^{+})\|$ < $\|E_{\text{obj}}(x_i) - E_{\text{scn}}(Y_i^{-})\|$ for all triplets in $\mathcal{T}$.
We learn the two mappings $E_{\text{obj}}$ and $E_{\text{scn}}$ with a single neural network. %

\begin{figure}[!t]
    \centering
    \includegraphics[width=0.48\textwidth]{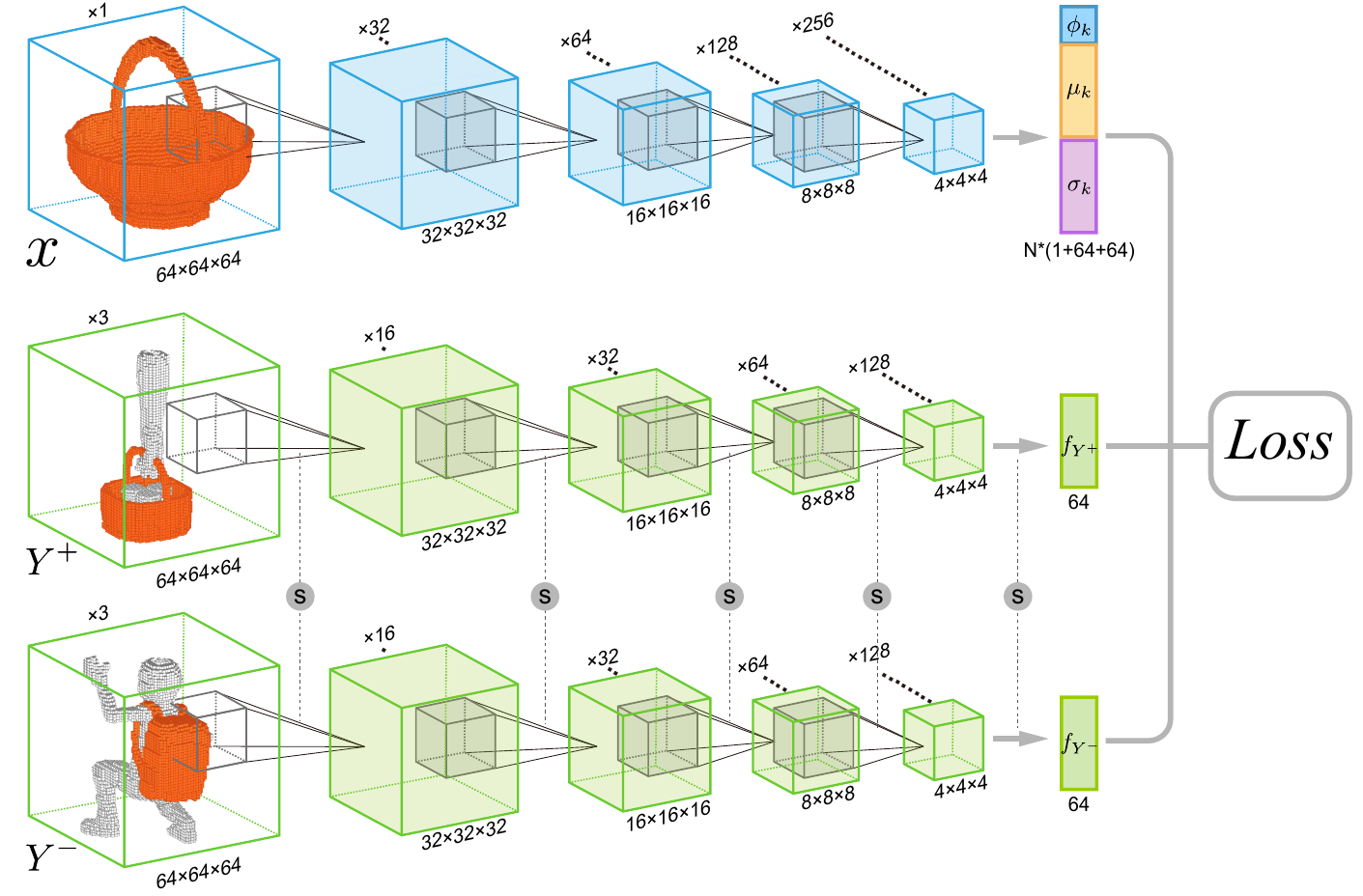}
\caption{The architecture of our functional similarity network -- fSIM-NET. The layers shown on the top row implement the $E_{\text{obj}}$ subnetwork, while the layers on the second and third rows implement the $E_{\text{scn}}$ subnetworks. We show over each volume the number of units of the same type that appear in the layer, while the dimensions of the data processed by each layer are written under the volume.}
\label{fig:network}
\end{figure}

\paragraph{Network architecture.}
We modify the original architecture of a triplet network to map different domains into the latent space.
Our network is composed of three subnetworks, as illustrated in Figure~\ref{fig:network}. The network takes as input one object and two scenes represented as 3D volumes composed of $64^3$ voxels. Two subnetworks implement $E_{\text{scn}}$ for the input \emph{scenes}, while one additional subnetwork implements $E_{\text{obj}}$ for isolated \emph{objects}. The subnetworks map their inputs to the interaction space, by converting the 3D volumes into feature vectors. The types and numbers of units that we use in each layer of the subnetwork are listed in the supplementary material.

Each $E_{\text{scn}}$ subnetwork is implemented with convolutional layers that map an input scene $Y$ into a $64$-dimensional vector $f_Y$, representing the coordinates of the central object of the scene in the interaction context space. The two $E_{\text{scn}}$ subnetworks share parameters as they compute the same function.

The $E_{\text{obj}}$ subnetwork maps an isolated object to the interaction context space. If every object in the world had a single functionality, then the mapping could be modeled as a straightforward one-to-one embedding. However, in practice, an object can serve multiple functionalities, and often there is correlation in the functionality of distinct object categories~\cite{hu16}. Thus, inspired by the work of Sung et al.~\shortcite{sung17} on shape completion, we learn a probabilistic mapping to the latent space using a Gaussian Mixture Model (GMM). Using a GMM, we can compute the expectation that an input object $x$ functions as represented by a scene $Y$:
\begin{equation}
\mathcal{E}(x, Y) = -\log \sum_{k=1}^{N}
\phi_k(x)\,p(f_Y|\mu_k(x),\sigma_k(x)),
\label{eq:expectation}
\end{equation}
where $f_Y$ is the mapping of $Y$ computed with $E_{\text{scn}}$,
$p$ is modeled as a Gaussian distribution, $N$ is the number of components in the GMM, and $\{ \phi_k, \mu_k, \sigma_k \}$ are the parameters of the $k$-th Gaussian in the model: component weights, mean, and variance, respectively. Note that these parameters are functions learned by the $E_{\text{obj}}$ subnetwork, which implement a probabilistic version of the mapping function $E_{\text{obj}}$. The $E_{\text{obj}}$ subnetwork is implemented with convolutional layers attached to a final layer that provides the parameters of the GMM (details provided in the supplementary material).

\paragraph{Training the network.}
The training consists of optimizing the parameters of the subnetworks to minimize a contrastive loss on the training triplets $\mathcal{T}$. The loss is similar in spirit to that of Wang et al.~\shortcite{wang14} and Sung et al.~\shortcite{sung17}, but considers the mapping for two different domains as captured by the expectation $\mathcal{E}$ in Eq.~\ref{eq:expectation}:
\begin{equation}
L(\mathcal{T}) = \frac{1}{n} \sum_{i=1}^{n} L(x_i, Y_i^{+}, Y_i^{-}),
\quad\text{with}
\label{eq:full_loss}
\end{equation}
\begin{equation}
L(x,Y^{+},Y^{-}) = \max \{ 0, m + \mathcal{E}(x, Y^{+}) - \mathcal{E}(x, Y^{-}) \},
\label{eq:loss}
\end{equation}
where $n = |\mathcal{T}|$ is the number of triplets, and $m$ is a {\em gap parameter} that helps control the convergence of the optimization. If the difference between the negative expectation and positive expectation is less than $m$, then the triplet contributes to the gradient when optimizing the network parameters. Otherwise, if the gap is satisfied, no contribution is incurred by the triplet. The loss ensures that objects with similar functionality are kept close in the mapping space, and those with dissimilar functionality are kept apart. The network is trained with the {\em Adam optimizer}, where all the subnetworks are trained together with the loss in Eq.~\ref{eq:full_loss}.

\paragraph{Functionality predictions.}
Once the network is trained, we can use the $E_{\text{obj}}$ and $E_{\text{scn}}$ subnetworks to predict various functionalities.
We can compute functional differences between two scenes $Y_1, Y_2$ as $\|E_{\text{scn}}(Y_1) - E_{\text{scn}}(Y_2)\|_2$, and between two objects $x_1, x_2$ as some PDF difference between $E_{\text{obj}}(x_1)$ and $E_{\text{obj}}(x_2)$.
However, we can also compute $\mathcal{E}(x, Y)$ to estimate the distance of any input object $x$ to a scene $Y$ in the training data, which allows us to predict the most probable functionalities of a given object $x$.

\begin{figure}[!t]
    \centering
    \includegraphics[width=0.48\textwidth]{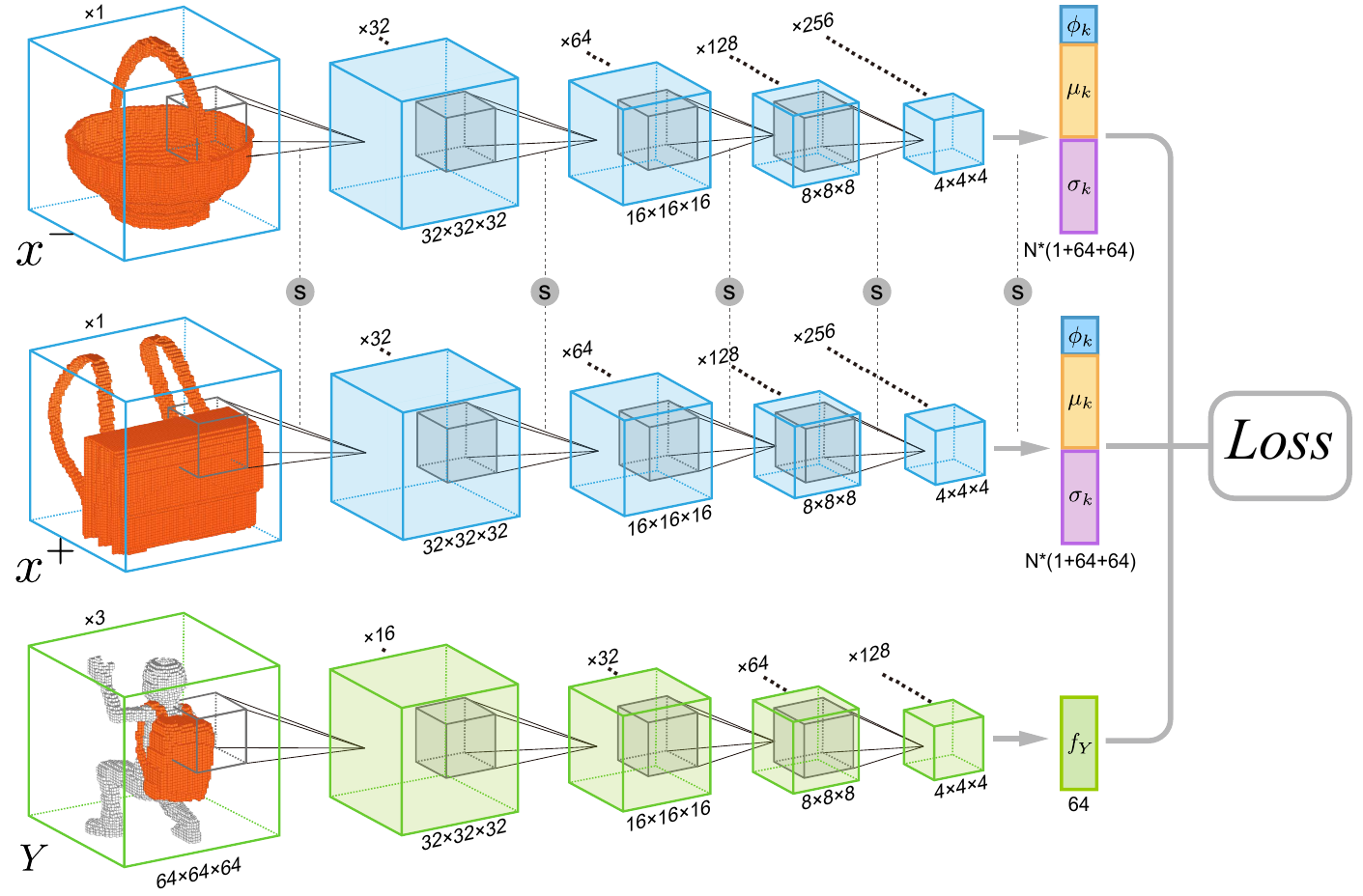}
\caption{Architecture of the fSIM-NET for learning the metric in the scene-to-object direction, composed of two $E_{\text{obj}}$ and one $E_{\text{scn}}$ subnetworks.}
\label{fig:directions}
\end{figure}

\begin{figure*}[!t]
    \centering
    \includegraphics[width=0.98\textwidth]{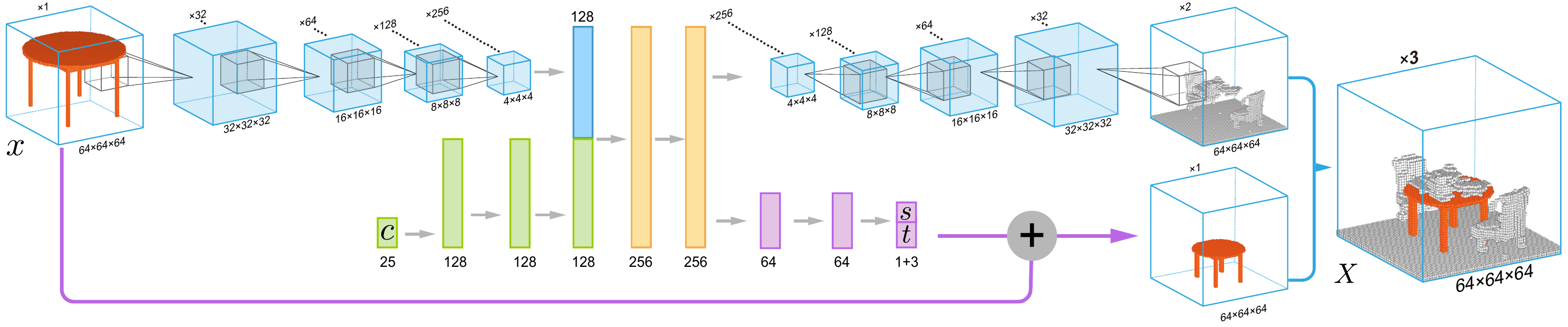}
\caption{The architecture of our interaction context generation network -- iGEN-NET. Given an input object $x$ (top-left) and functionality label $c$ (bottom-left), the network generates an output scene $X$ and places $x$ into this scene (right), based on transformation parameters $s$ and $t$.}
\label{fig:generative}
\end{figure*}

\paragraph{Scene-to-object distance.}
Through its training procedure, fSIM-NET learns a metric optimized for the {\em object-to-scene direction}. That is, using the latent space learned, the distances from an object $x$ to two scenes $Y_i$ and $Y_j$ are comparable, since the triplets used for training constrain such relative comparisons. However, the metric is not explicitly optimized for the {\em scene-to-object direction}. That is, given two objects $x_i$ and $x_j$, their distances to a scene $Y$ are not necessarily comparable. Such comparisons can be used to find objects that best support a given functionality $Y$, for example, if we want to replace the central object in the scene $Y$ with a different object $x$ that fulfills the same functionality as the central object. 

However, using the same ideas as in the fSIM-NET, we can build a network composed of two $E_{\text{obj}}$ (that share parameters) and one $E_{\text{scn}}$ subnetworks, to obtain a metric  in the scene-to-object direction (Figure~\ref{fig:directions}). To train the network, we require a set of suitable triplets $\mathcal{T}'$ which include distances involving one scene $Y$ and two objects $x^+$, with the same, and $x^-$, with different, functionality as the central object in $Y$. Then, our goal can be posed as learning $E_{\text{obj}}$ and $E_{\text{scn}}$ so that $\|E_{\text{obj}}(x^+) - E_{\text{scn}}(Y)\|$ < $\|E_{\text{obj}}(x^-) - E_{\text{scn}}(Y)\|$, for all the triplets in $\mathcal{T'}$. The new network is then trained with a loss similar to Eq.~\ref{eq:loss}, adapted to $\mathcal{T}'$:
\begin{equation}
L(x^{+},x^{-},Y) = \max \{ 0, m + \mathcal{E}(x^{+}, Y) - \mathcal{E}(x^{-}, Y) \},
\end{equation}
where the expectations are computed with the GMMs learned by each $E_{\text{obj}}$ subnetwork and the $E_{\text{scn}}$ subnetwork.

\paragraph{Classification networks.}

Our network architectures can be adapted specifically for classification purposes.  These can be used in applications where we are  interested in classifying an input object or scene into one or more functionality categories. In this context, the training triplets are provided with classification labels of the functionality category of both objects and scenes. To create a classification network for the object-to-scene direction, we add two fully-connected layers at the end of the $E_{\text{obj}}$ subnetwork, and learn a function $L(x)$ that translates the output parameters of the GMM into class  probabilities. Similarly, to perform classification in the scene-to-object direction, we add two fully-connected layers at the end of a $E_{\text{scn}}$ subnetwork, and learn a labeling function $L(Y)$.

\begin{figure*}[!t]
    \centering
    \includegraphics[width=0.98\textwidth]{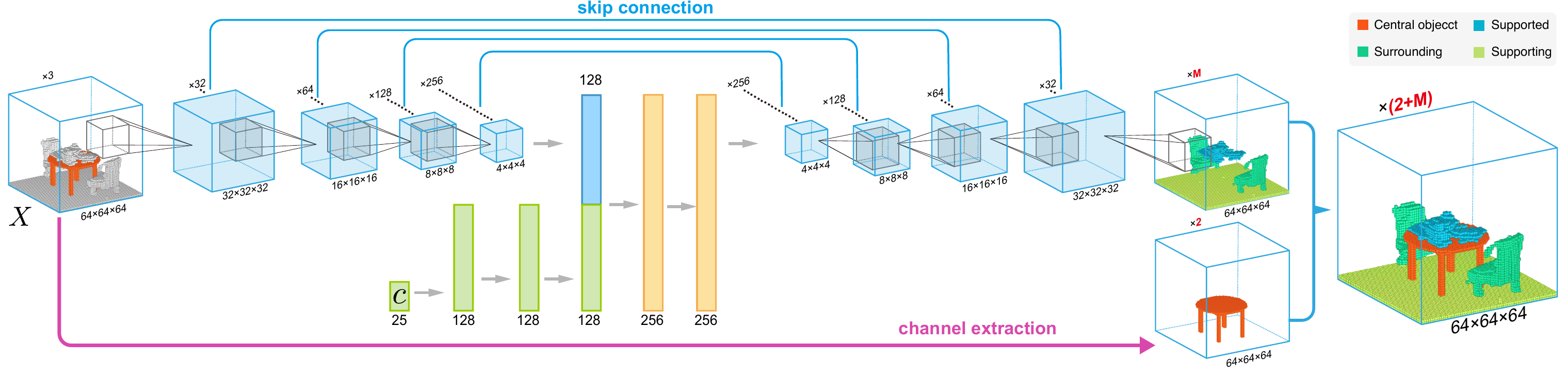}
\caption{The architecture of our segmentation network -- iSEG-NET. Given an input scene $X$ (top-left) and functionality label $c$ (bottom-left), the network segments the interacting objects in the scene into different groups. The central object, extracted from the input encoding, can then be recombined with the output segmented scene (right). 
        }
\label{fig:segmentation}
\end{figure*}

\section{iGEN-NET: Context generation network}

Going beyond inferring or classifying functionality, we introduce iGEN-NET, which is a network capable of {\em generating}
a context scene for an object given in isolation. The synthesized scene is composed of objects interacting with the input object in ways that demonstrate its functionality. The network takes as input an individual voxelized object $x$ and a functionality label $c$ represented as a one-hot vector, and outputs a voxelized scene $Y$. The label can be predefined or selected among the high probability labels predicted by classifying $x$ as described in the previous section. Although objects can support more than one functionality, by providing a single label as input, we define the specific functionality that we wish to illustrate within the scene $Y$ generated by iGEN-NET.  

The interaction context generation is accomplished with three subnetworks, as shown in Figure~\ref{fig:generative}. The object is first embedded into a feature space with the use of a convolutional subnetwork, providing a 128-dimensional embedding vector $e_x$ for the object. Fully-connected layers combine the object's embedding vector $e_x$ and functionality label $c$ to provide a 256-dimensional feature vector.  Lastly, a {\em decoder} subnetwork takes this feature vector as input and synthesizes the output interaction context scene. 
In parallel, a {\em spatial transformer network}~\cite{jaderberg15} composed of a few fully-connected layers takes as input the same feature vector and defines the scaling and translation needed to place the input object $x$ into the scene. Detailed information regarding the layers of each subnetwork is given in the supplementary material.

By combining the output of the synthesis and placement subnetworks, we obtain a complete scene $Y$ with the input object $x$ placed in the appropriate location, and synthesized objects as the interaction context for the central object $x$. 
As we show in Section~\ref{sec:results}, by varying the input label $c$, we are able to generate various scenes that illustrate different functionalities of the same object.

To train the network, we provide examples of voxelized scenes with a central object and multiple interacting objects, together with the functionality label. We define two loss functions: the loss function for the synthesis subnetwork is the mean of the voxel-wise cross-entropy between the training and synthesized interaction contexts (not including the central object), while the loss for the placement subnetwork is the sum of the L2 norms for the scaling values and translation vectors of the central object between the training and synthesized scenes. We first train the placement subnetwork alone. Then, we fix the parameters of this subnetwork and train the synthesis subnetwork with the voxel cross-entropy loss~\cite{girdhar16}. Finally, we fine-tune the entire network with a loss function that is the sum of the two losses of the subnetworks.

\section{iSEG-NET: Segmentation network}

The output of the generation network is a voxelized scene including voxels of three types: the central object, context, and empty voxels. The goal of iSEG-NET is to separate the context voxels into objects with different interaction types with the central object.
Assuming that there are $M$ different types of interactions involved in our dataset (e.g., chairs are placed next to a table, and books on top of a table), we use the network to find the probability of each of the context voxels to be labeled as one the $M$ possible interactions. 

The network takes as input a context scene, that includes only the context and empty voxels of the output of iGEN-NET, and outputs a vector of size $M$ of probabilities $p_i^j$ for every voxel $j$ in the context voxels, where $p_i^j$ is the probability of voxel $j$ being labeled as interaction type $i$.
Similarly to iGEN-NET, the network is composed of encoder and decoder convolutional subnetworks with skip connections, illustrated in Figure~\ref{fig:segmentation}. The encoder reduces the input volume into a 128-dimensional feature vector, which is concatenated with the output of fully-connected layers that process the label of the scene. The concatenated feature vector is further processed and decoded to yield a volume with the probability of the context voxels. 

To train the network, we prepare training data by labeling each scene with the interaction type of each interacting object. In our work, we consider $M=18$ interaction labels, which represent all the interaction types that we observed in our dataset. These include interactions such as {\em supported}, {\em supporting}, {\em sitting}, {\em riding}, {\em hanging}, and {\em typing}, etc.
The loss function for the segmentation network is defined as the mean of the voxel-wise cross-entropy between the ground-truth and predicted labeling.

\begin{figure}[!t]
    \centering
    \includegraphics[width=0.48\textwidth]{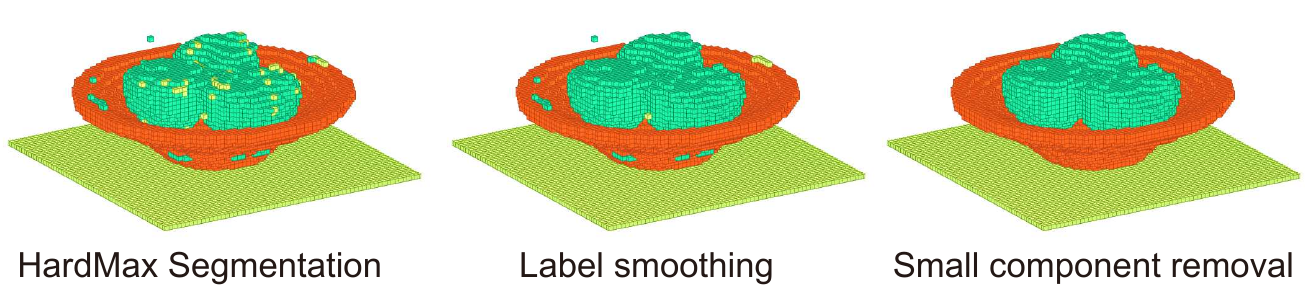}
\caption{Segmentation result of an interaction context. Left: using just a hard maximum. Middle: after smoothing the labels given by the network with graph cut. Right: after removing small connected components in the volume.}
\label{fig:refinement}
\end{figure}

To create the final segmentation, we can simply take the maximum of the probabilities of each voxel. However, the generated results in this case may be noisy and contain small floating parts; see Figure~\ref{fig:refinement}. To smooth the labeling of the voxels, we solve a multi-label optimization problem by applying graph-cuts to the probability distributions~\cite{boykov2004experimental}. We build a graph where  each voxel of an interacting object is a node, and there is an edge connecting any two neighboring voxels, using 26 connectivity. The data cost for a voxel and label $l$ is set to $1-p_l$, where $p_l$ is the probability that the voxel is of label $l$. For the smoothness term between different labels, we compute the frequency of each pair of labels being neighbors in the training data and set the cost to be $1-f_{i,j}$, where $f_{i,j}$ is the normalized frequency that the labels $l_i$ and $l_j$ are neighbors. 
Once we have a labeling for each voxel, we find all the connected components in the volume for any label. If the component size is smaller than $10\%$ of the size of the maximal component of its label, we remove the component.

Lastly, by combining back the voxels of the central object, we obtain a voxelized scene $Y$ that contains multiple component objects having different interactions with the central object.  
The voxelized scene $Y$ can be further refined by retrieving high-resolution models to substitute the synthesized objects in the scene.

Note that, since the scenes generated with the iGEN-NET are represented using $64^3$ voxels, there is no guarantee that each object in the scene is complete and isolated. Thus, to address this limitation, we also benefit from introducing the segmentation network to label the voxels, so that individual objects can be retrieved to constitute a meaningful scene, described as follows.

\begin{figure}[!t]
    \centering
    \includegraphics[width=0.48\textwidth]{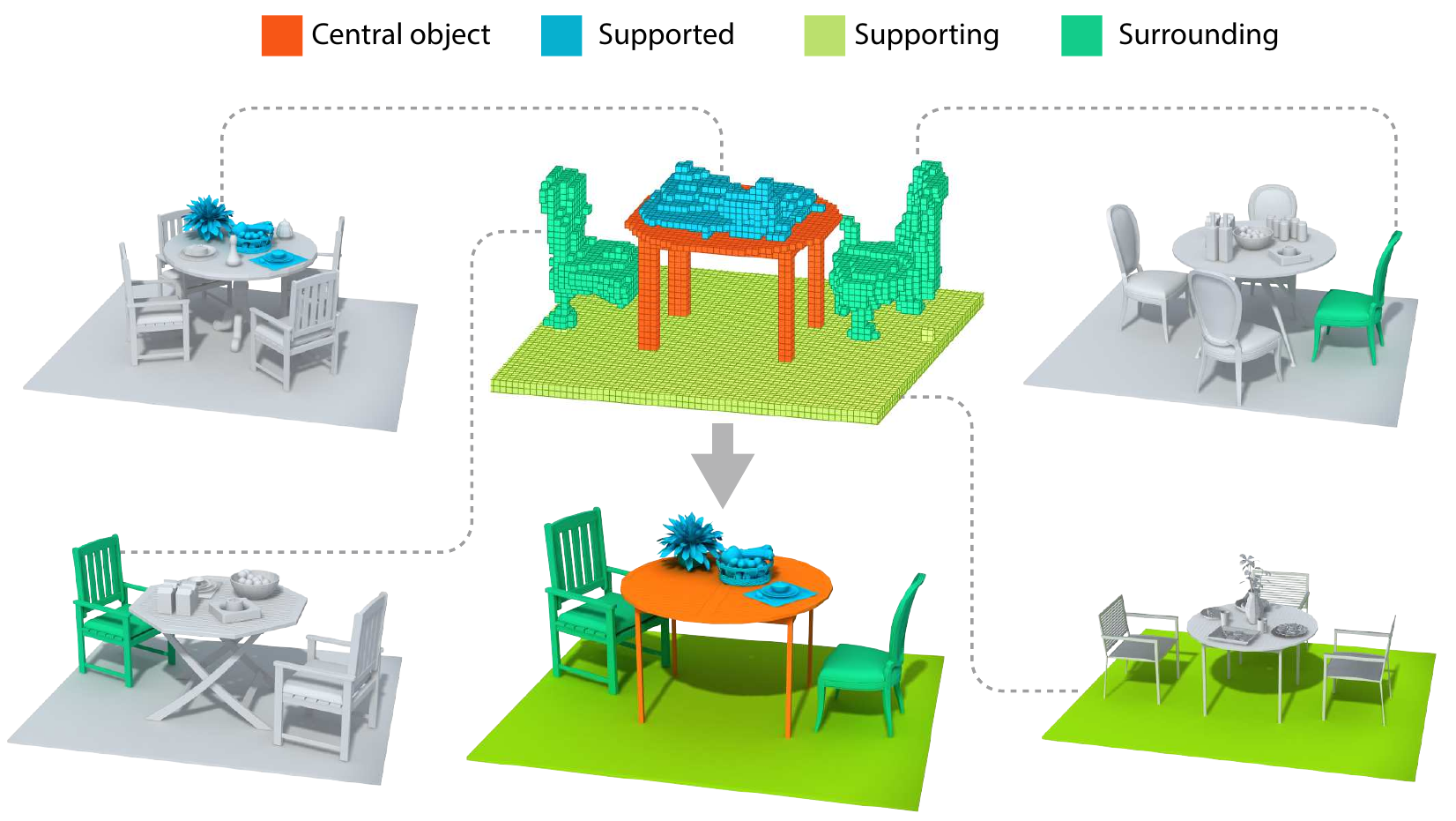}
\caption{Scene refinement: given a synthesized, segmented scene (top-center), we retrieve higher resolution models from scenes in our dataset (left and right), to replace connected components of voxels with the same interaction type. The result is a refined scene with more detailed models (bottom-center).}
\label{fig:mesh_fitting}
\end{figure}

\paragraph{Scene refinement.} 
To replace the segmented voxels in a scene with 3D objects, we retrieve
objects from the scenes in our dataset which are the most similar to
each connected component in the segmented voxels (Figure~\ref{fig:mesh_fitting}). 
To define a good similarity measure for retrieval, we train a classification network to map each object in our dataset to its labeled interaction type, and then use the last feature layer of this classifier to encode both the objects and segmented voxels in our generated scenes. The L2 distance of this feature vector is used for retrieval. Once all the objects that will replace the segments are retrieved, we scale and translate them to place them around the central object, so that the position and size of their bounding boxes relative to the bounding box of the central object is similar to the relation between the bounding boxes of their corresponding voxels and the central object in the generated scene.

\section{Evaluation and results}
\label{sec:results}

In this section, we present results obtained with the networks
that we introduce in this work, and further evaluate them with comparisons to
previous works. Additional evaluation experiments and timing information are provided in the supplementary material.

\paragraph{Dataset.} We use a dataset of central objects and their surrounding scenes derived from two sources. The first source is the dataset of Hu et al.~\shortcite{hu16}, composed of 15 classes of objects. Each central object is given in the context of a scene demonstrating its functionality. There are 1-5  interactions for each central object, where the types of interactions included in the dataset can be inferred from the geometry of the objects. We further extended this dataset with additional categories from the ModelNet40 dataset~\cite{wu15}. We added the 10 categories that do not overlap with the dataset of Hu et al.~\shortcite{hu16} and which have functionalities that can be derived from the objects' geometries. We randomly selected 40 objects from each category and complemented them with surrounding scenes. In total, our dataset contains 1,008 scenes. The category names and number of scenes are listed in the supplementary material. 

Note that some of the scenes in our dataset were extracted from larger
scenes taken from datasets of previous works, while other scenes were
created by an artist, where the objects in
the scenes were collected from various sources. Thus, we cannot fully
guarantee that similar pieces of objects do not repeat in some of the scenes,
although we ensured sufficient diversity in the arrangements of objects.
The affordance labels assigned to segmented scenes are {\em Carrying}, {\em Contained}, {\em Hanging}, {\em Holding}, {\em Hung}, {\em In front}, {\em Lighted}, {\em Lying}, {\em On side}, {\em Overhanging}, {\em Pushing}, {\em Riding}, {\em Side-supporting}, {\em Sitting}, {\em Supported}, {\em Supporting}, {\em Surrounding} and {\em Typing}. Examples of scenes with these labels are shown in the supplementary material.

\begin{figure}[!t]
    \centering
    \includegraphics[width=0.48\textwidth]{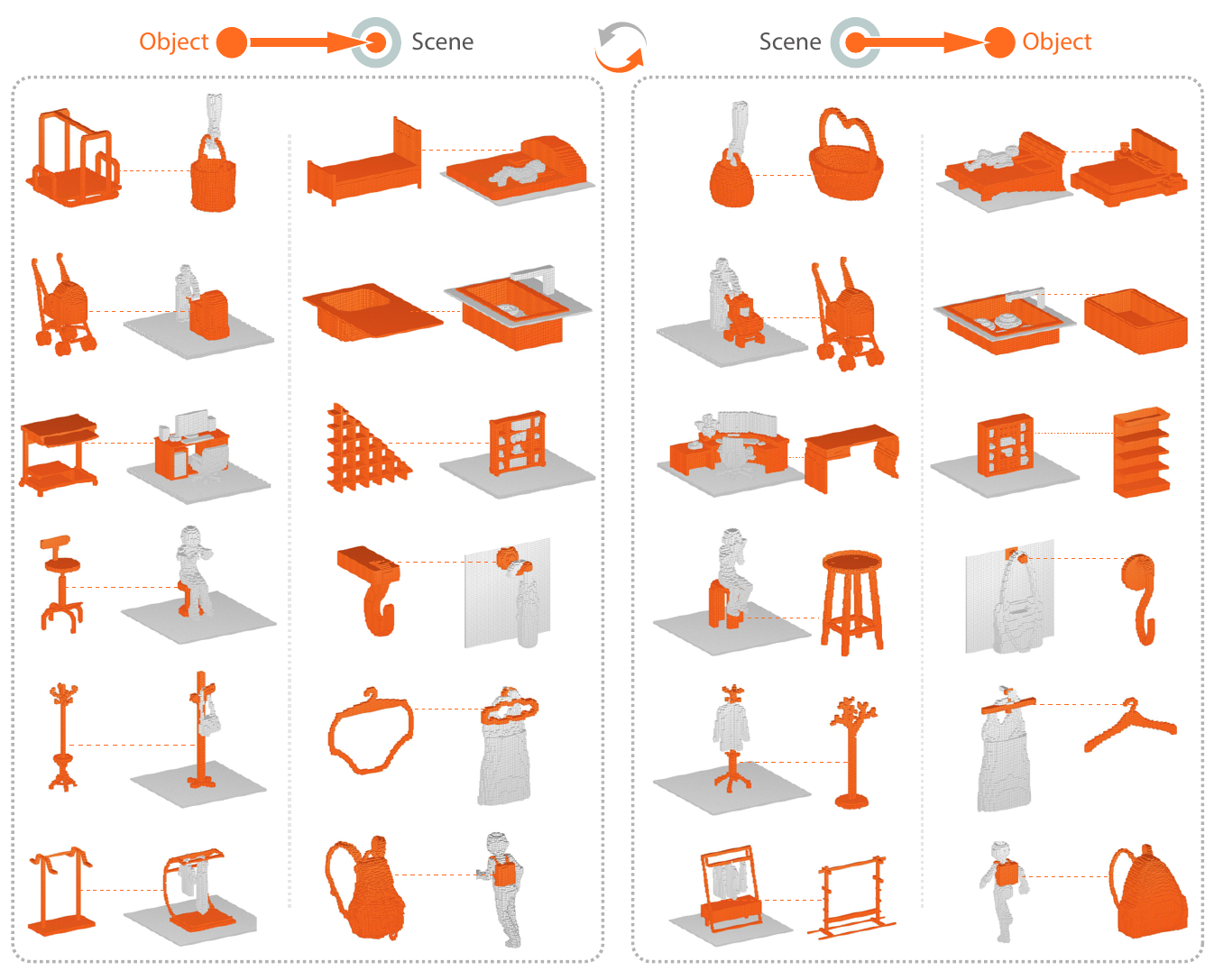}
\caption{Selected results showing fSIM-NET used for object and scene retrieval. Left: Retrieved scenes with functionalities most similar to query objects. Right: Retrieved objects most similar to query scenes.}
\label{fig:gallery}
\end{figure}

\paragraph{Functional similarity network.} We first evaluate our deep neural network fSIM-NET
that estimates the functional similarity between objects and scenes, in the two
possible directions: object-to-scene, and scene-to-object.

\paragraph{Data collection for metric learning.} 
The triplets needed for training depend on the direction of the distance measure that we are training. For the object-to-scene direction, the triplets are of the form $(x_i, Y_i^{+}, Y_i^{-})$, while for the scene-to-object direction, the triplets are $(x_i^{+}, x_i^{-}, Y_i)$.
As our dataset contains category labels of the scenes, we use these to define our training examples. Using a pair of scenes in the same category, we can build a positive example pair $(x_i^{+}, Y_i)$ or $(x_i, Y_i^{+})$, by extracting the central object from one scene to define $x_i^{+}$ or $x_i$. To add negative examples to the pairs and form triplets, we randomly sample one scene from a different category and take either the whole scene to create $Y_i^{-}$ or the central object for $x_i^{-}$.

\begin{figure}[!t]
    \centering
    \includegraphics[width=0.48\textwidth]{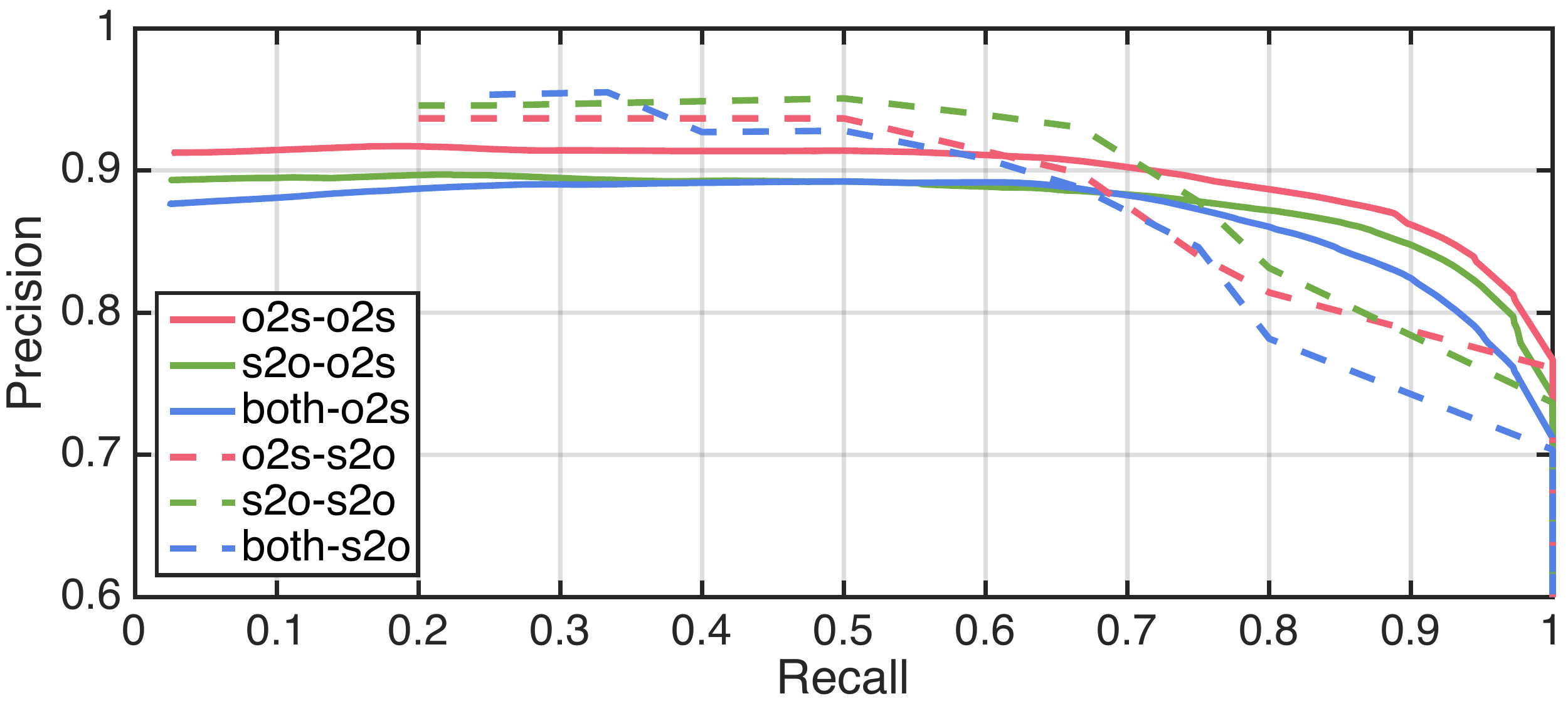}
\caption{Performance of our functional similarity network in a quantitative evaluation. Items in the legend are in the form ``training direction-test direction'', where o2s denotes ``object-to-scene'' and s2o is ``scene-to-object''.}
\label{fig:bi-direction}
\end{figure}

\paragraph{Results and evaluation.}
Figure~\ref{fig:gallery} shows selected results of using fSIM-NET for object and scene retrieval, according to the two possible directions of the similarity measure. We note that all of the top retrieved objects and scenes are meaningful examples of the query's functionality, and include a variety of human-object and object-object interactions, such as sitting, hanging, and support. 

In Figure~\ref{fig:bi-direction}, we present a quantitative evaluation of the performance of the network in terms of precision/recall, where a result is considered correct if the query and retrieved result are positive examples of each other. For these experiments, we perform a cross-validation evaluation where we divide the dataset into a 9:1 training to test ratio. We evaluate the effect of training and testing the network in each possible direction. In general, we observe that the network provides a precision around 0.9 for recall rates up to 0.7. We also see that the network provides the best results when trained in the specific direction that is being evaluated. However, the precision is comparable when we train and test the network in opposite directions. Thus, in practice, training the network in one direction tends to also constrain the other direction of the metric.
Interestingly enough, we found that when training the network in both directions together, the performance is slightly worse than training the network in each direction alone. A possible reason could be that the margin in both directions constrains the embedding too much.

In the supplementary material, we present results on pose invariance for additional evaluation of the fSIM-NET. We also examine its generalization capabilities, although our experiments are not fully conclusive due to the size and nature of our training set.

\paragraph{Comparison to alternative approaches.} We compare our similarity network to alternative approaches on the scene-to-object and scene-to-scene directions as follows. We do not provide comparisons on the object-to-scene direction since, to the best of our knowledge, there are no previous works that optimize a metric in this direction.

\begin{figure}[!t]
    \centering
    \includegraphics[width=0.48\textwidth]{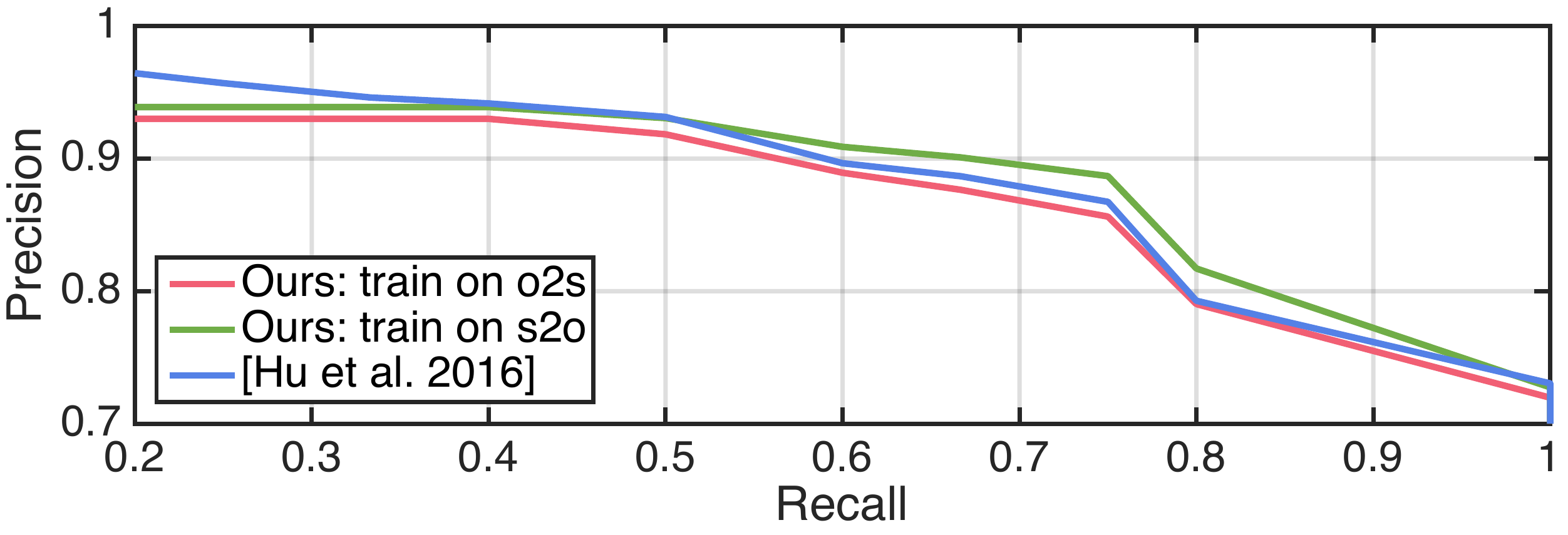}
\caption{Comparison of our functional similarity network to the functionality model of Hu et al.~\shortcite{hu16}. Please refer to the text for details.}
\label{fig:comparison_to_ICON2}
\end{figure}

\begin{figure}[!t]
    \centering
    \includegraphics[width=0.48\textwidth]{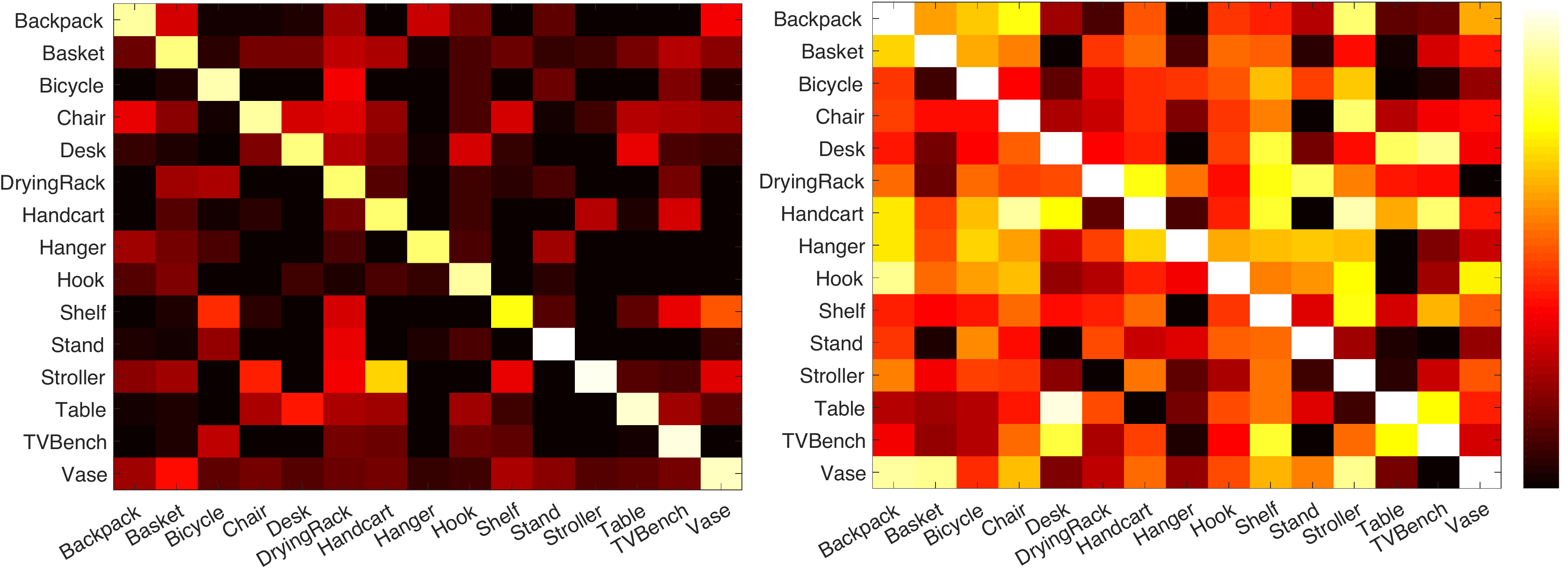}
		\caption{Comparison of correlations between categories estimated by our
		method (left) to those estimated by the method of Hu et al.~\shortcite{hu16} (right). }%
		\label{fig:comparison_to_ICON2_corr}
\end{figure}

\paragraph{Scene-to-object direction.}
In Figure~\ref{fig:comparison_to_ICON2}, we compare our method to the work of Hu et al.~\shortcite{hu16}, which learns a model of functionality with the handcrafted interaction context (ICON) descriptor. Their model can be used for similarity assessment mainly in the scene-to-object direction, as descriptor distances are constrained to be comparable only in this direction. 
We observe that our method trained on the scene-to-object direction obtains results comparable to those of Hu et al., which shows that the geometric features extracted by their method capture the essential functionality features, where the use of a simple learning method provides good results. The advantage of our method is that the entire mapping is trained end-to-end, without requiring complex geometric pre-processing.
Figure~\ref{fig:comparison_to_ICON2_corr} provides a summary of the comparison by presenting matrices that display the correlation between all pairs of categories in our dataset. To compute an entry (i, j) of this matrix, the average distance of all shapes in class $i$ to those in class $j$ is computed. The inverse of these averages are shown by color mapping in the matrix. Thus, larger values (closer to white) imply higher correlation. Note how our network provides a much clearer indication of correlation. For example, strollers and handcarts are strongly correlated, but strollers and backpacks are not.

\paragraph{Scene-to-scene direction.}
We compare our work to two alternative approaches: (i) The original ICON descriptor of Hu et al.~\shortcite{hu15}, which is suitable for comparisons in the scene-to-scene direction, since the descriptor can only be built from an input scene. (ii) Siamese and Triplet networks. A common approach for learning a distance measure between entities of the same type is to train Siamese or Triplet networks to map the entities to a latent space, where a distance metric can be defined~\cite{wang14}. Thus, we train such networks to test alternative approaches to our fSIM-NET. 

To train Siamese networks, the training data is in the form of positive or negative scene pairs, while for Triplet networks, the training data is in the form of triplets containing two positive examples and one negative example. Scenes in our dataset are classified into different functionality categories. Thus, for defining training pairs for the Siamese network, any two scenes in the same category are considered as positive pairs, while any two scenes in different categories are considered as negative pairs. For defining training triplets for the Triplet network, we combine each positive scene pair with a randomly selected scene from each different category, which constitute negative examples.

\begin{figure}[!t]
    \centering
    \includegraphics[width=0.48\textwidth]{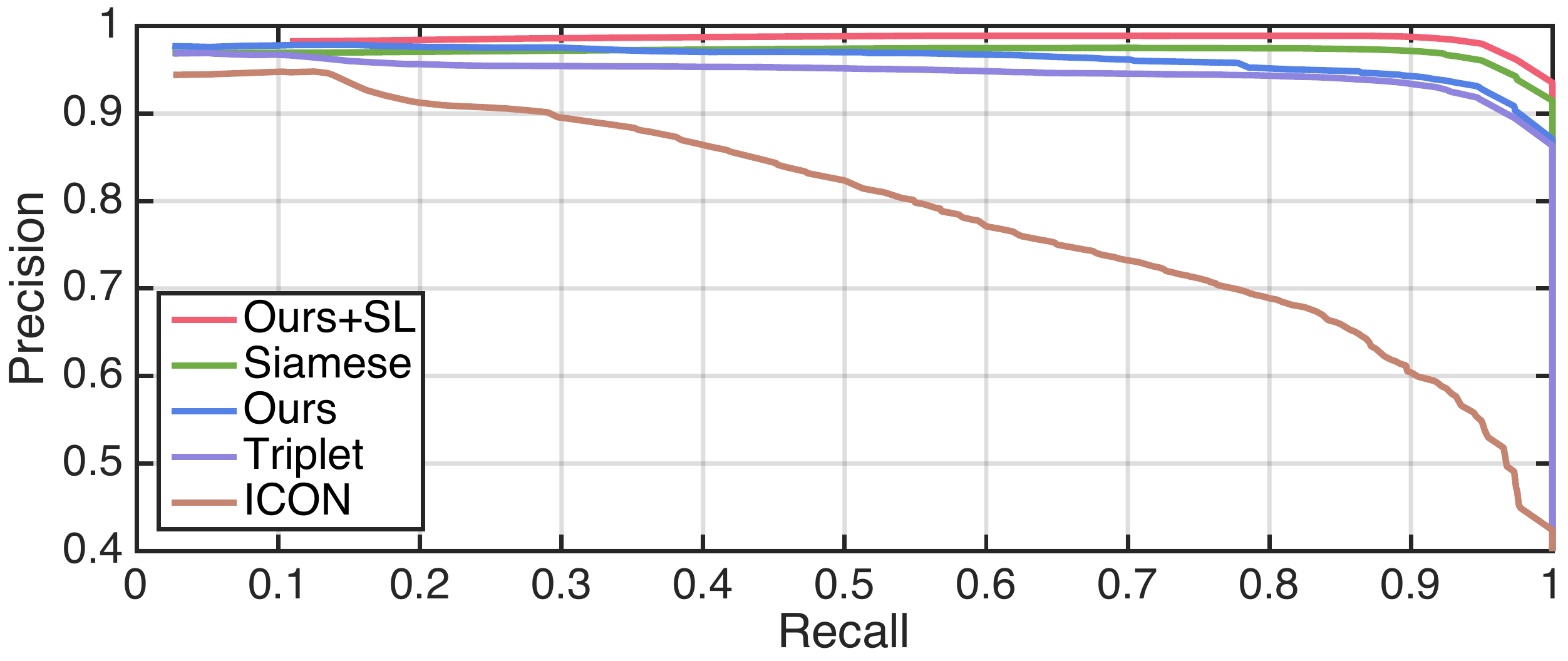}
\caption{Comparison of our functional similarity network (denoted ``Ours'') to two alternative networks (``Siamese'' and ``Triplet'') and the ICON descriptor, for measuring scene-to-scene distances. ``Similarity + SL'' denotes a version of our network trained with the Siamese Loss (SL). }
\label{fig:comparison_to_ICON1}
\end{figure}

Note that our fSIM-NET is not designed for directly providing a scene-to-scene distance measure, but an object-to-scene distance. However, we can use the $E_{scn}$ subnetwork to derive a scene-to-scene distance measure. Moreover, we evaluate an alternative version of our network which is trained by adding the same loss and training data of the Siamese network to the $E_{scn}$ subnetwork. We compare these two versions of our network to the alternatives in a cross validation evaluation scheme with a 9:1 training to test ratio. The results are shown in Figure~\ref{fig:comparison_to_ICON1}. 

First, we observe that all neural networks obtain much higher precision than the ICON descriptor. Second, we observe that our similarity network trained together with the Siamese loss obtains the best result. As expected, our network trained without the Siamese loss obtains a slightly lower precision due to two reasons. First, the network does not use all the training data in our dataset, since we sample triplets for each positive pair by adding a random negative example from another class, which may not involve all possible pairs of objects. Second, the network uses a loss that does not directly optimize scene-to-scene distances, as the Siamese network does.

\paragraph{Classification.}
To evaluate the classification version of our network, we assign to each central object the ground-truth functionality label of its corresponding scene. Next, we compute the classification accuracy simply as a binary value indicating whether the correct label was predicted by our network or not, averaging this value for all test shapes.
We compare the classification accuracy for three different alternatives: (i) We add two fully connected layers after the last layer of the $E_{obj}$ subnetwork in our fSIM-NET, and train our network together with a classifier. (ii) We use the object-to-scene distance provided by the fSIM-NET to find the nearest neighbor scene to a query object, assigning the label of the scene to the object. (iii) We use a direct neural network classifier that uses exactly the same network architecture as the $E_{obj}$ subnetwork in (i), but is not trained together with the $E_{scn}$ subnetwork.

In a cross-validation experiment with a 9:1 train to test ratio over 25 object categories, we observe that the direct network, i.e., alternative (iii), obtains an accuracy of 84\%, the nearest neighbor approach, i.e., alternative (ii), obtains 92\%, and the fSIM-NET extended with a classifier, i.e., alternative (i), performs slightly better at 94\%. Thus, the use of our network provides results that are around 10\% higher than when using a direct classifier, with a slight improvement when attaching a classifier to the network.

\begin{figure}[!t]
    \centering
    \includegraphics[width=0.48\textwidth]{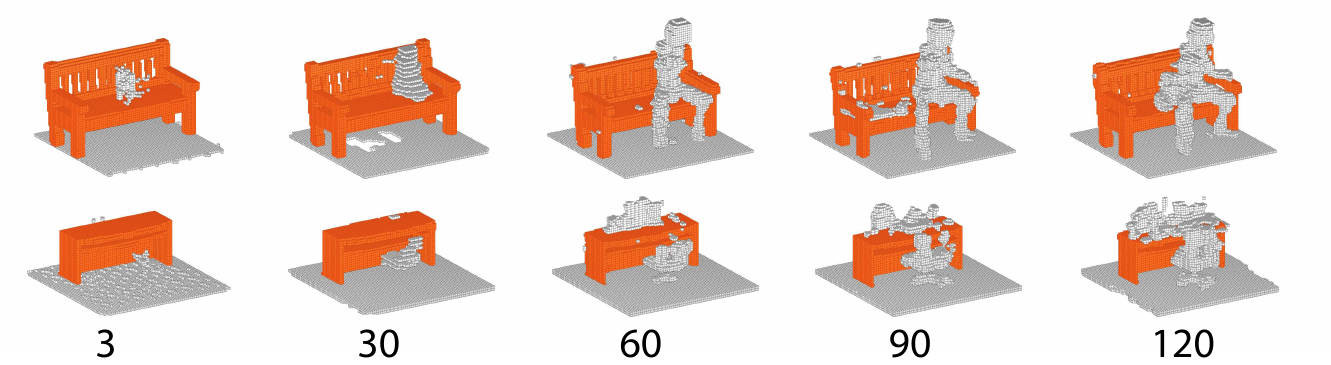}
\caption{Interaction contexts (in gray) synthesized for two central objects (in orange) with networks trained with the indicated number of epochs.}
\label{fig:progressive_gen}
\end{figure}

\paragraph{Interaction context generation network.} To train the generation
network, we use $90\%$ of the scenes and their central objects in our dataset
as training data, along with their labels, and test on the remaining $10\%$. Note that a single category label  is assigned to each scene, e.g., desk, table, bed, etc.

We present results of using our iGEN-NET to synthesize interaction contexts, and analyze how the network adapts to different types of input. Figure~\ref{fig:progressive_gen} illustrates the learning progress of the generative network, where we show the synthesis results for a same testing shape obtained with networks at different training stages. We observe that after 120 epochs of training, the network is able to synthesize interaction contexts effectively, where the scenes contain a rich variety of details and meaningful object shapes, such as the human sitting on the bench or objects on top of the table.

\begin{figure}[!t]
    \centering
    \includegraphics[width=0.48\textwidth]{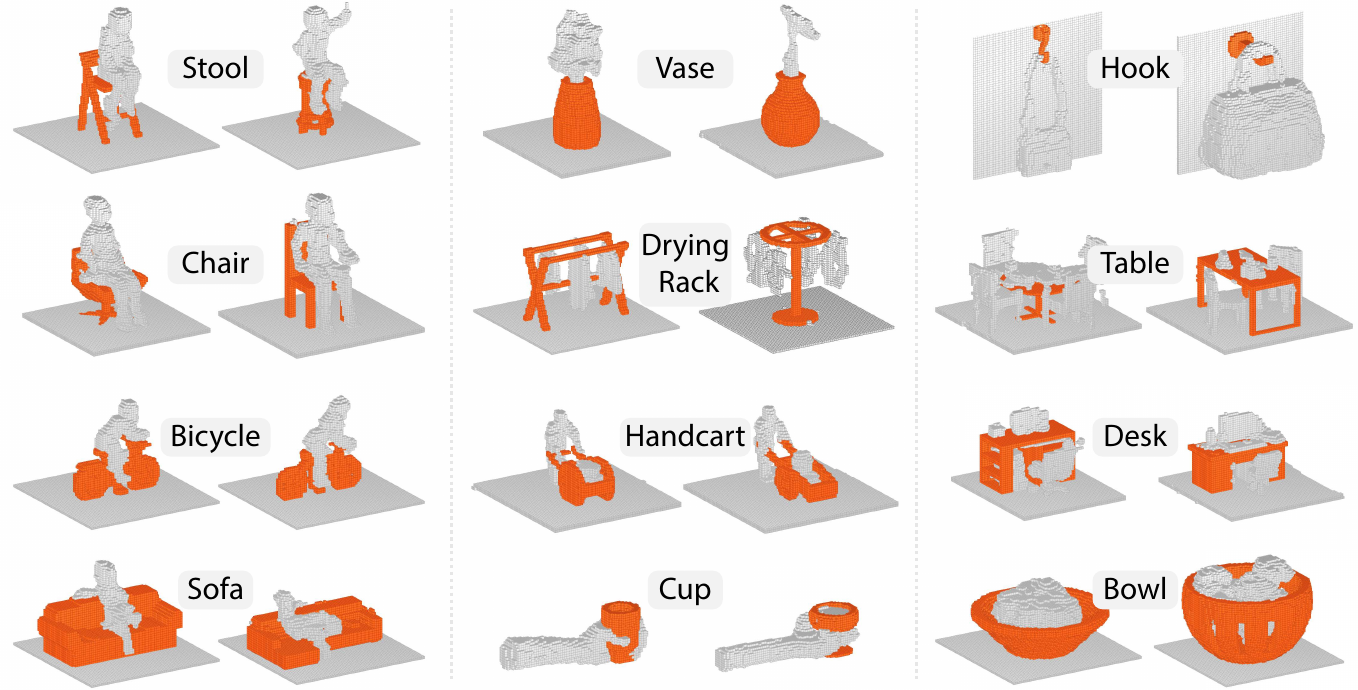}
\caption{Gallery of interaction context generation results. We show two
selected central objects from each category (in orange), and the scenes
generated by our network iGEN-NET around the objects (in gray). Note how the
scenes generated for each pair of objects are quite distinct and adapt
to the geometry of the objects, e.g., drying racks and cups.}
\label{fig:gallery_gen}
\end{figure}

Figure~\ref{fig:gallery_gen} presents a gallery of interaction context
synthesis results, where we show two different objects from the same
class and their synthesized contexts. In these examples, we observe how
the results of the network properly adapt to the geometry of the input shape. For example, hanging clothes are synthesized on the appropriate regions of the drying racks, even though one rack is straight and the other is circular, and a synthesized hand grasps a cup differently depending on whether a handle is present or not.

\begin{figure}[!t]
    \centering
    \includegraphics[width=0.48\textwidth]{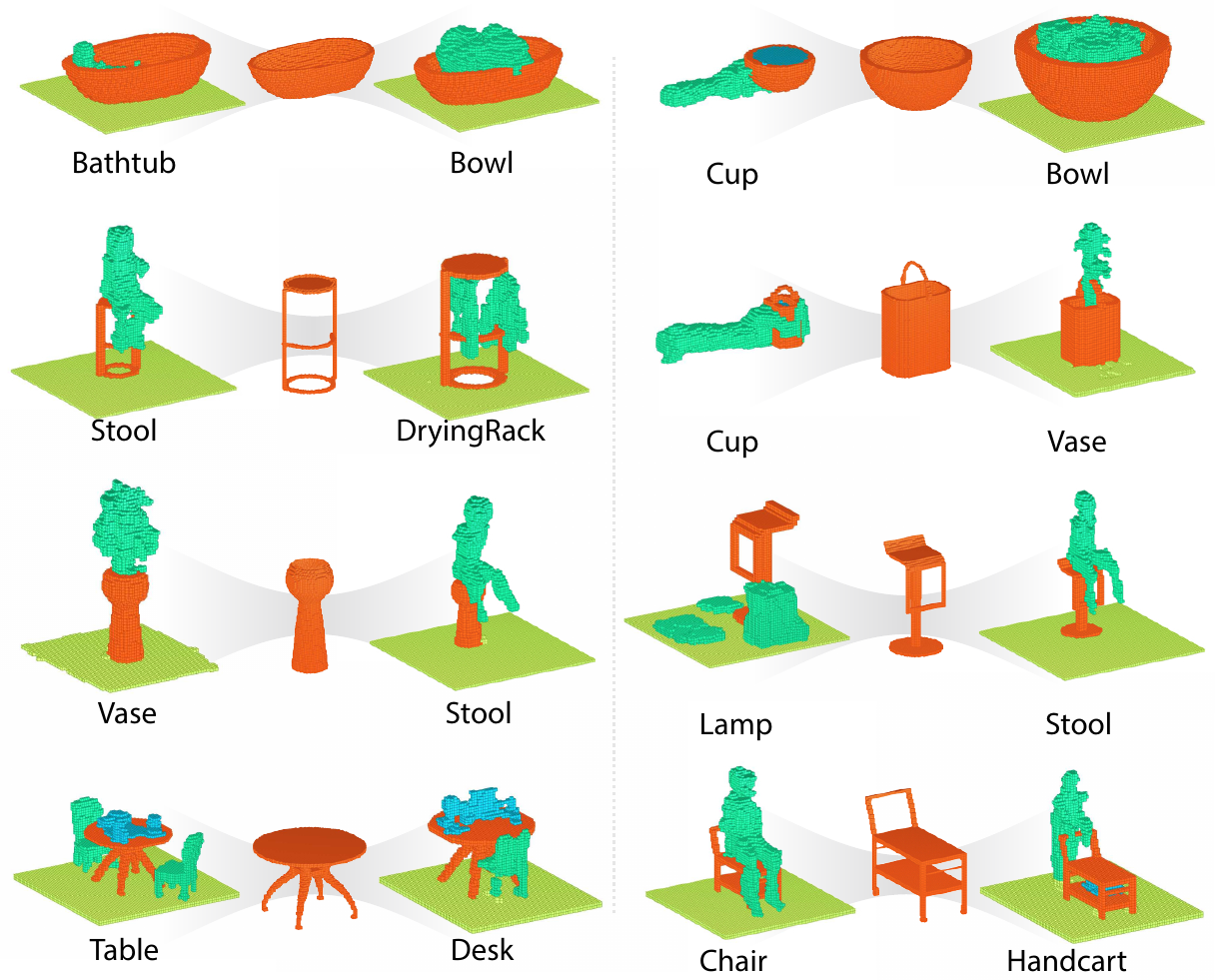}
\caption{Gallery of multi-functionality synthesis results with their segmentations. Given the object in the middle, we generate the two interaction contexts on the left and right, based on the labels denoted below. Note how the generated scenes adapt to both the input object and label provided.}
\label{fig:multi_func}
\end{figure}

In addition, Figure~\ref{fig:multi_func} shows a gallery of synthesis results for shapes that can serve more than one functionality. In these examples, we synthesize two interaction contexts for the same input shape while specifying a different functionality label. We observe that the synthesized contexts adapt satisfactorily to the label provided. For example, a table can easily function as a table or desk. However, in each case, the synthesized scene is different in the types of objects placed on the table, showing the subtle difference between tables and desks. Similarly, a basket can also function as a vase or cup, as shown by the synthesized interaction contexts. The examples for the chair and handcart show that, although handcarts have wheels attached to them, their geometry in fact approximates well the functionality of a chair. In summary, we observe in these qualitative examples how the synthesized interaction contexts adapt to both the geometry of the input objects and the functionality label provided.

\begin{figure}[!t]
    \centering
    \includegraphics[width=0.48\textwidth]{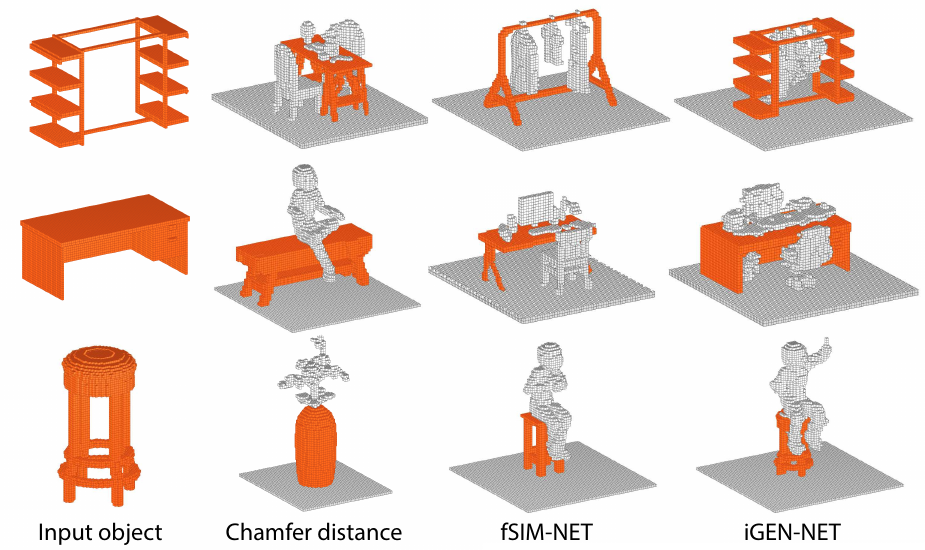}
\caption{Comparison of scenes synthesized with our iGEN-NET to retrieval-based alternatives: The most similar central object retrieved using Chamfer distance (shown with its corresponding scene) and the most similar scene retrieved with the fSIM-NET.}
\label{fig:ret_vs_gen}
\end{figure}

\paragraph{Comparison to alternative synthesis approaches.}  
We compare the results obtained with our iGEN-NET to two retrieval
methods that could serve as alternative baselines for generating interaction contexts. The first baseline involves retrieving a central object from the dataset most similar to the query comparing only isolated objects. Specifically, we use the Chamfer distance which can be used to compare two voxelized objects~\cite{su17}. 
The second baseline involves retrieving the closest scene to the query object with our fSIM-NET. Next, we could take the objects from the retrieved scenes and place them around the query object to generate an interaction context. Note that, in our comparison, we do not explicitly perform this transfer of surrounding objects, but show the retrieved scenes to demonstrate how difficult it would be to generate scenes with these baseline methods. 

In Figure~\ref{fig:ret_vs_gen}, we show three examples that are representative of
the results in a large experiment. We observe that the first baseline often retrieves scenes that have completely different functionality than the query, and would thus lead to incorrectly synthesized scenes. The second baseline can retrieve objects that are slightly different from the query, e.g., a drying rack with a central bar rather than two bars. Thus, transferring the objects in a straightforward manner would lead to the context scene not properly adapting to the query object, e.g., floating clothes, while the iGEN-NET generates results adapted to the geometry of the query object. 

Finally, as discussed in Section~\ref{sec:related}, scene synthesis methods in the literature that take functionality into consideration typically model only human-object interactions~\cite{savva16,ma16}. For pairs of objects, mainly co-occurrence is considered, but not object-object interactions as in our iGEN-NET.

\paragraph{Diversity of generated scenes.}
To evaluate the diversity of the synthesized output, we perform a comparison of the variation in the training data compared to the variation in the synthesized data. Ideally, we would compare the generated scenes to the training data with a similarity measure such as the Chamfer distance. However, this would not provide conclusive evidence as the output in general only partially overlaps with the training data. Thus, to evaluate the diversity of generated data, we first compute the Chamfer distance between each pair of training scenes to obtain a mean and variation of their similarity. Then, we compute the mean and variation for the generated scenes. The Chamfer distance mean and variation for the training set are 7.96 and 37.19, respectively, while those for the generated set are 7.40 and 28.13.

Moreover, for each training scene, we find the most similar generated scene and compute their Chamfer distance. We then compute the average of these distances for the entire dataset. The mean and variation of the distances from each training scene to its closest generated scene are 1.3775 and 0.0013, respectively, implying that we can find a generated scene that is close enough to each training scene, considering the distance mean and variation of the training scenes as reported above. This experiment indicates that the diversity of the output is close to that of the training data.

\begin{figure}[!t]
    \centering
    \includegraphics[width=0.48\textwidth]{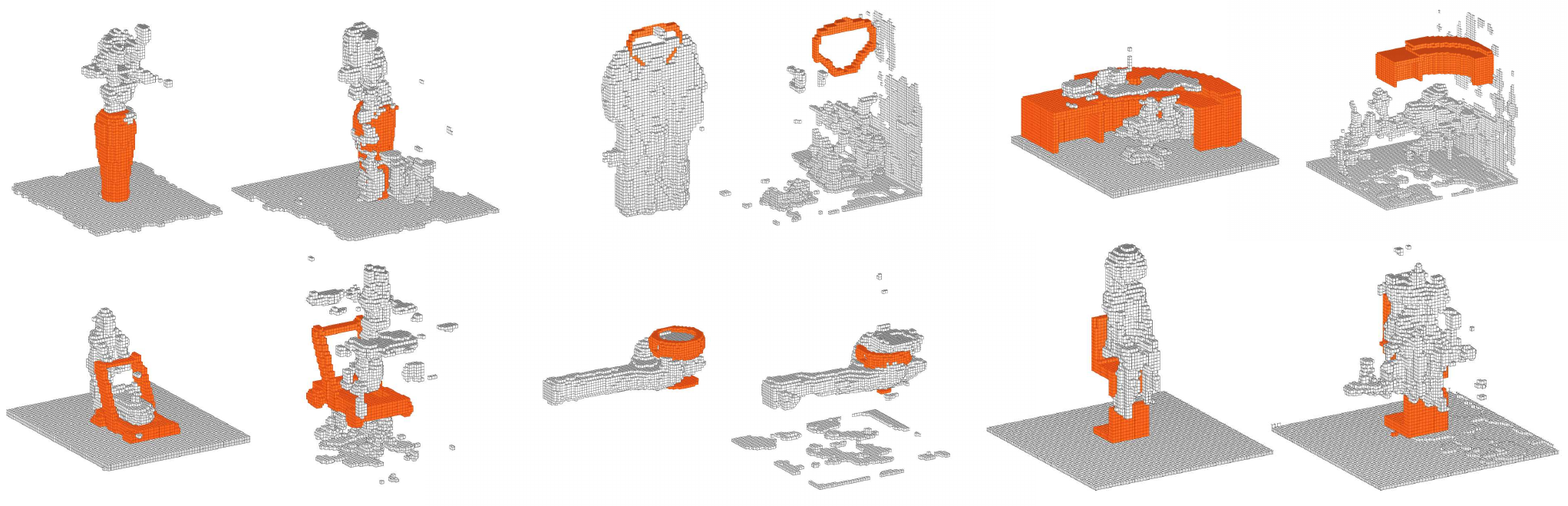}
\caption{Comparison of interaction contexts generated with (left) and without (right) giving the functional label of the central object. Note the noise in the generated scenes and how different functionalities get mixed up.}
\label{fig:compare_label}
\end{figure}

\begin{figure}[!t]
    \centering
    \includegraphics[width=0.48\textwidth]{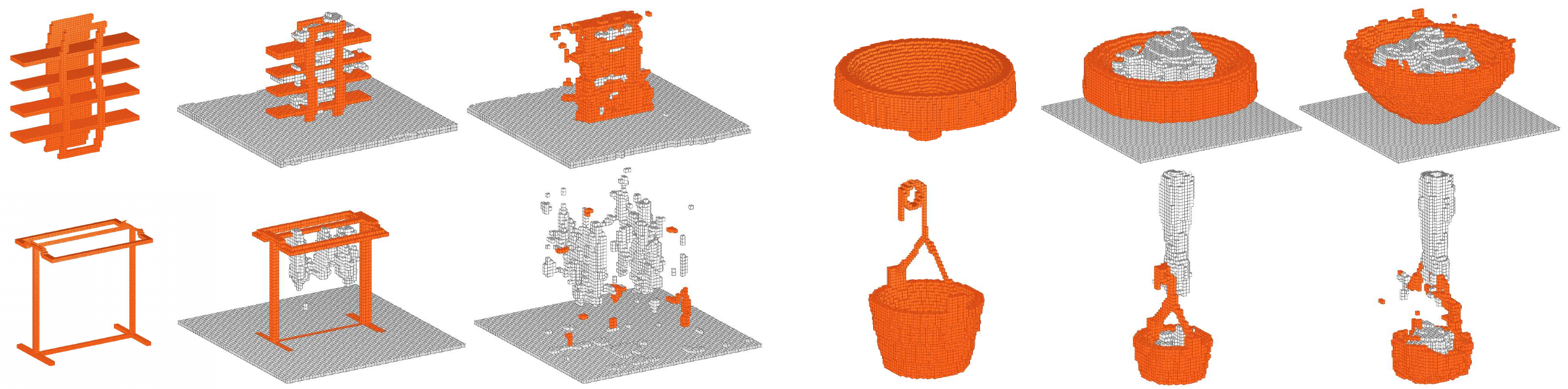}
\caption{Comparison of interaction contexts generated with (left) and without (right) the transformer
subnetwork. Note how the generated scenes display noise and structural
problems.}
\label{fig:compare_transformer}
\end{figure}

Although the synthesized contexts adapt to the provided object and label, our network is a non-stochastic regressor and thus it cannot provide different outputs when given the same object and label. As indicated by the diversity experiment above, the network does not gurantee to synthesize novel interacting objects, but positions the objects existing in the training data so that they appropriately adapt to the given object.
Although novel objects are not generated, our network can synthesize object usage scenarios that do not exist in the training data, e.g., there is no sitting scenario for the handcart shown in the bottom right of Figure~\ref{fig:multi_func}. Moreover, the label provided to the network ensures that artifacts in the synthesized scenes are minimized, in contrast to scenes generated without this information, as shown in Figure~\ref{fig:compare_label}. Moreover, the transformer subnetwork also contributes to the quality of the generated scenes by keeping but also properly scaling and placing the input object, in contrast to the results generated without this sub-network, as shown in  Figure~\ref{fig:compare_transformer}. Without the transformer subnetwork, the network tends to generate the most common or average scene in each category. %

\paragraph{Segmentation network.} 
To train the segmentation network iSEG-NET, we manually segment all the scenes in
our dataset into separate objects, and assign a common label to all the objects that have similar interactions with the central object, e.g., all the books on a shelf receive the same label of ``supported''. The segmented volumes are used as training data for the network. We provide  examples of our labeling in the supplementary material.

\begin{figure}[!t]
    \centering
    \includegraphics[width=0.48\textwidth]{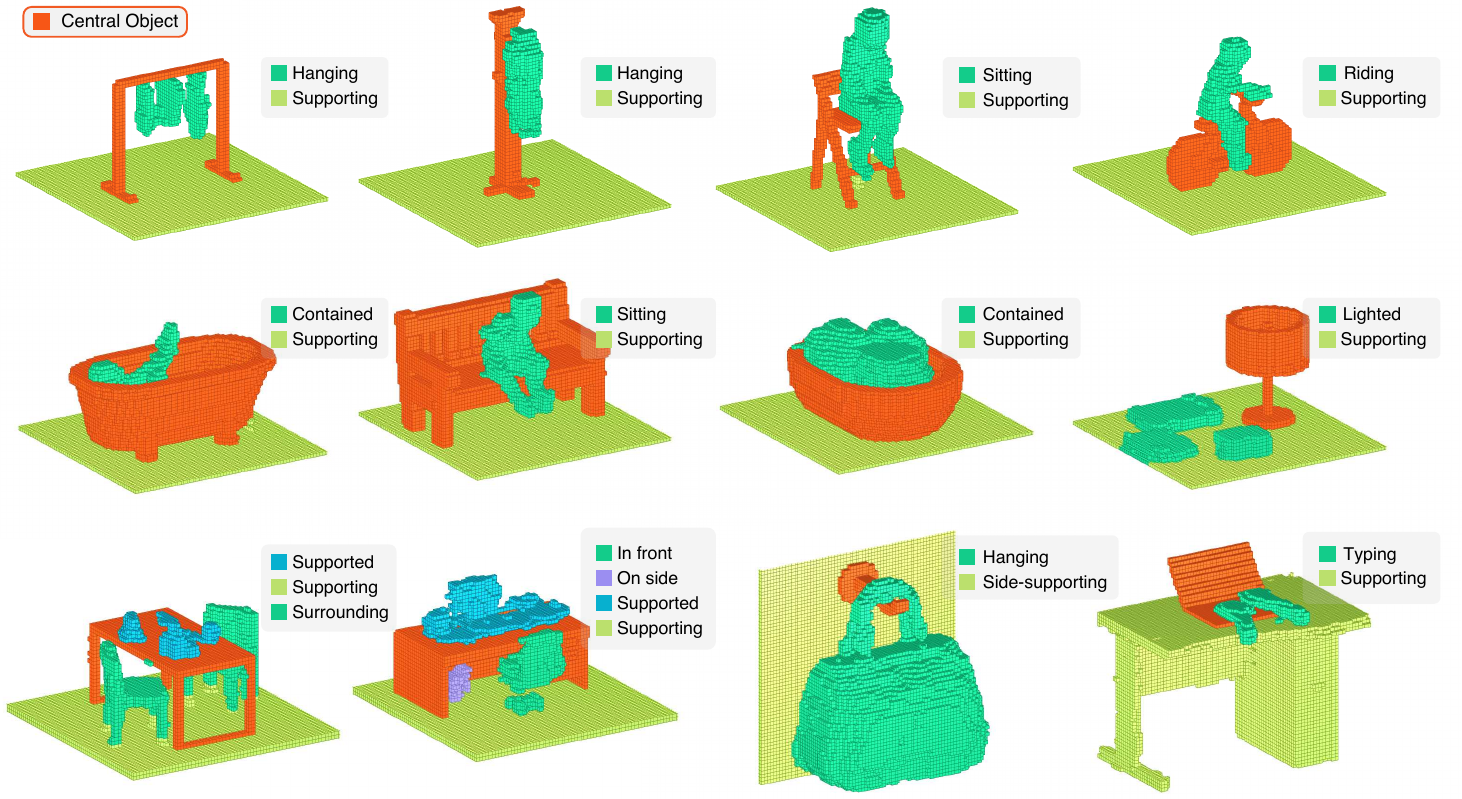}
\caption{Gallery of scene segmentation results obtained with our segmentation network. The various interaction labels are shown in different colors, while the central objects are colored orange.}
\label{fig:gallery_seg}
\end{figure}

Figure~\ref{fig:gallery_seg} shows a gallery of segmented interaction
contexts obtained with our iSEG-NET. In addition, Figure~\ref{fig:multi_func} shows the results on scenes generated for objects with multiple functionality. We observe in all of these examples that the network is able to segment objects into groups that have similar interactions with the central object, and identify the correct labels for the groups. For example, the scene for a desk is segmented into the chair besides the desk, the floor which supports the desk, and one group for all the objects that are supported by the desk. In general, the segmentations include between one to five different groups of objects. To provide a quantitative evaluation of the iSEG-NET, we compute the segmentation accuracy of the network in a cross-validation experiment, finding that the average accuracy for a segmentation obtained by applying a hard maximum to the label probabilities is 98\%.

\begin{figure}[!t]
    \centering
    \includegraphics[width=0.48\textwidth]{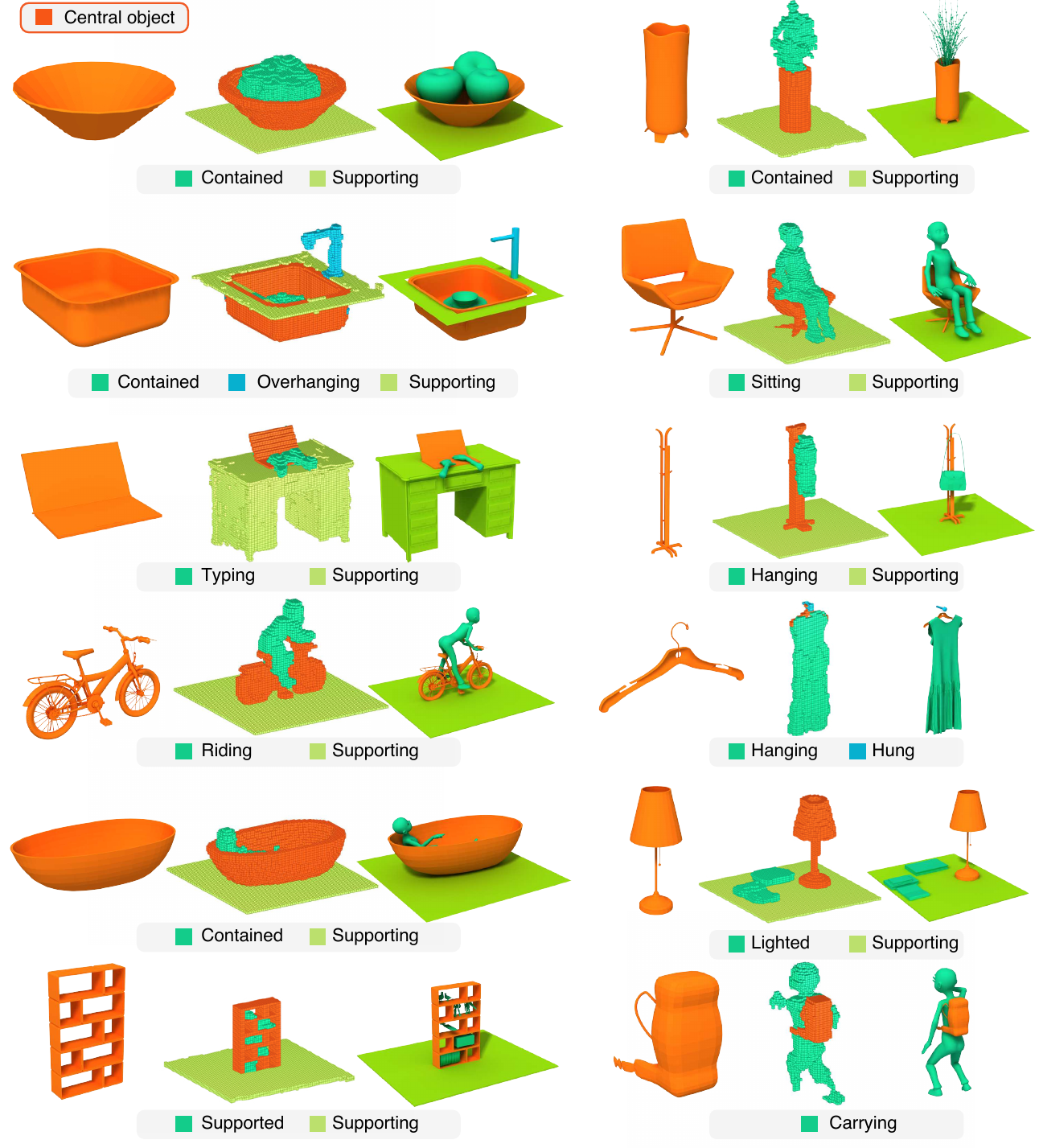}
\caption{Gallery of scene refinement results. In each example, we show the input shape, generated scene, and refined scene, including interaction labels.} 
\label{fig:gallery_mesh}
\end{figure}

\paragraph{Scene refinement.}
Figure~\ref{fig:gallery_mesh} shows examples of refining the synthesized scenes by replacing sets of voxels with higher-resolution models retrieved from a dataset of objects. We observe how the segmentation into interaction types allows the post-processing to select meaningful objects to compose the scenes, and place them in appropriate positions and orientations.

\section{Discussion, limitations, and future work}
\label{sec:future}

In this work, we enable functional understanding and functionality hallucination of isolated 3D objects with the introduction of three deep neural networks: fSIM-NET, iGEN-NET, and iSEG-NET. Specifically, the networks allow not only to predict the functionality of an object, but also to substantiate it by generating an example scene that demonstrates how the object interacts with surrounding objects to reveal a functional usage scenario. 

We show in our evaluation that fSIM-NET outperforms handcrafted descriptors and models for functionality prediction proposed in previous works. In addition, the scenes exemplifying functional uses of objects generated by our iGEN-NET incorporate both human-object and object-object interactions, and adapt to the geometry of the objects.
Finally, the iSEG-NET segments the synthesized output so that it can be more easily analyzed and refined, e.g., by replacing voxels with higher-resolution meshes.

As a first step in functional analysis using deep neural networks, our work has several limitations that can suggest interesting directions for future work. To start, we utilized the category labels of scenes in our dataset to select positive and negative examples for creating the training triplets of the fSIM-NET. This can limit the potential of the network in discovering cross-category functionalities, e.g., between desks and tables, if desks are added as negative examples of tables and vice-versa. A possible direction for improving the learned similarity measure is to directly collect observations on the similarity of triplets, e.g., via crowdsourcing. In this manner, the training examples would potentially also capture natural correlations that exist between different categories.

Currently, the interaction contexts generated by the iGEN-NET are quite limited in terms of scene complexity, as they mainly demonstrate the functionality of one object. It would be interesting to extend this approach to generate larger and more complex scenes that display broad functionalities, e.g., a living room or kitchen. Furthermore, our iSEG-NET segments groups of objects in the synthesized scenes according to their interaction types, which is the natural grouping for interaction contexts. However, post-processing methods would also benefit from a segmentation of the scene into individual objects, enabling the refinement of each individual object. 
In addition, our proposed scene refinement method allows us to adequately exchange voxels for meshes in many of the scenes. However, there are different possibilities for improving this simple refinement method. One option would be to incorporate semantic constraints specific to each interaction type, e.g., two objects with ``supported'' and ``supporting'' interactions should be in contact. 

Finally, the interactions that can be handled by our method are limited
to static functionalities that can be inferred from the geometry of
objects. A few recent works also model {\em dynamic} interactions for analyzing the functionality of objects. For example, Hu et al.~\shortcite{hu17} represent the mobility of shape parts with a linear model involving only two static part configurations, while Pirk et al.~\shortcite{pirk17} encode the dynamic use of objects by tracking the trajectory of particles on the surface of an object during an interaction. Incorporating part mobility or dynamic trajectories into our functional analysis framework would certainly extend the range of functionalities that can be predicted and demonstrated with synthesized scenes.

\section*{Acknowledgements}
We thank the anonymous reviewers for their valuable comments. This work was supported in part by NSFC (61602311, 61522213, 61761146002, 61861130365), 973 Program (2015CB352501), GD Science and Technology Program (2015A030312015), Shenzhen Innovation Program (JCYJ20170302153208613, JCYJ20151015151249564) and NSERC Canada (611370, 611649, 2015-05407).

\bibliographystyle{ACM-Reference-Format}
\bibliography{iconNetMain}

\end{document}